%% file: paper.tex
\documentclass[]{bytedance_seed}

\usepackage[toc,page,header]{appendix}

\usepackage{tikz}
\usepackage[export]{adjustbox} 
\usepackage{minitoc}

\usepackage{placeins}   

\usepackage{longtable}
\usepackage{booktabs}
\usepackage[utf8]{inputenc} 
\usepackage[T1]{fontenc}    
\usepackage{hyperref}       
\usepackage{url}            
\usepackage{booktabs}       
\usepackage{amsfonts}       
\usepackage{nicefrac}       
\usepackage{microtype}      
\usepackage{xcolor}         
\usepackage{tablefootnote}
\usepackage{caption}
\usepackage{wrapfig}  

\usepackage{tikz}
\usepackage{graphicx}
\usepackage{booktabs}
\input{sec/math_commands.tex}

\title{Chain-of-Action: Trajectory Autoregressive Modeling for Robotic Manipulation}

\author[1,2,5,*]{Wenbo Zhang}
\author[3]{Tianrun Hu}
\author[2]{Yanyuan Qiao}
\author[3]{Hanbo Zhang}
\author[4]{Yuchu Qin}
\author[5]{Yang Li}
\author[5]{Jiajun Liu}
\author[1]{Tao Kong}
\author[2,\dagger]{Lingqiao Liu}
\author[1,\dagger]{Xiao Ma}

\affiliation[1]{ByteDance Seed}
\affiliation[2]{The University of Adelaide}
\affiliation[3]{NUS}
\affiliation[4]{CAS}
\affiliation[5]{CSIRO}

\contribution[*]{Work done at ByteDance Seed}
\contribution[\dagger]{Corresponding authors}

\abstract{We present \textbf{Chain-of-Action (CoA)}, a novel visuo-motor policy paradigm built upon \emph{Trajectory Autoregressive Modeling}. Unlike conventional approaches that predict next step action(s) \textit{forward}, CoA generates an entire trajectory by explicit \textit{backward} reasoning with task-specific goals through an action-level Chain-of-Thought (CoT) process. This process is unified within a single autoregressive structure: (1) the first token corresponds to a stable keyframe action that encodes the task-specific goals;  and (2) subsequent action tokens are generated autoregressively, conditioned on the initial keyframe and previously predicted actions.
This backward action reasoning enforces a global-to-local structure, allowing each local action to be tightly constrained by the final goal. To further realize the action reasoning structure, CoA incorporates four complementary designs: continuous action token representation; dynamic stopping for variable-length trajectory generation; reverse temporal ensemble; and multi-token prediction to balance action chunk modeling  with global structure. As a result, CoA gives strong spatial generalization capabilities while preserving the flexibility and simplicity of a visuo-motor policy. Empirically, we observe CoA achieves the state-of-the-art performance across 60 RLBench tasks and 8 real-world manipulation tasks.}

\date{June 11, 2025}

\checkdata[Project Page]{\url{https://chain-of-action.github.io}}

\correspondence{\email{xiao.ma@bytedance.com}, \email{lingqiao.liu@adelaide.edu.au}}

\begin{document}
\maketitle

\vspace{-2pt}
\section{Introduction} \label{sec:intro}
\vspace{-4pt}

\input{sec/1.intro}

\vspace{-2pt}
\section{Related work} \label{sec:related}

\vspace{-2pt}
\input{sec/2.related}

\section{Chain-of-Action for robotic manipulation} 
\label{sec:method}

\input{sec/3.2.method}

\section{Implement details} 
\label{sec:Implement}
\input{sec/3.3.impl}

\section{Experiments} \label{sec:exp}
\vspace{-2pt}

\input{sec/5.exp}

\section{Conclusion}
\input{sec/6.conc}

\bibliographystyle{plainnat}
\bibliography{main}

\clearpage

\beginappendix

\input{sec/7.appendix}

\end{document}

%% file: sec/math_commands.tex
\usepackage{amsmath,amsfonts,bm}

\usepackage{pifont}

\def\eqref#1{(\ref{#1})}

\def\1{\bm{1}}

\DeclareMathAlphabet{\mathsfit}{\encodingdefault}{\sfdefault}{m}{sl}
\SetMathAlphabet{\mathsfit}{bold}{\encodingdefault}{\sfdefault}{bx}{n}

\usepackage{graphicx}
\usepackage{url}            
\usepackage{booktabs}       
\usepackage{nicefrac}       
\usepackage{microtype}      
\usepackage{wrapfig}

\usepackage{enumitem}

\usepackage[ruled,noend,linesnumbered]{algorithm2e}

\usepackage{multirow}
\usepackage{tablefootnote}

\usepackage{color}
\usepackage{colortbl}
\usepackage{xcolor}

\definecolor{blgrey}{rgb}{0.6,0.6,0.6}
\definecolor{bblue}{rgb}{0.855,0.933,0.98}
\definecolor{dblue}{HTML}{5297D6}
\definecolor{gainred}{rgb}{0.1,0.5,0.3}
\definecolor{citecolor}{HTML}{0071BC}
\definecolor{linkcolor}{HTML}{ED1C24}

\usepackage{listings}
\usepackage{color}

\definecolor{dkcyan}{cmyk}{1,0,0,.25}
\definecolor{dkgreen}{rgb}{0,0.6,0}
\definecolor{gray}{rgb}{0.5,0.5,0.5}
\definecolor{mauve}{rgb}{0.58,0,0.82}

\def\smallcaption{\small}

\newcommand\para[1]{\noindent\textbf{#1}\,}



%% file: sec/1.intro.tex
Visuo-motor policies have made substantial progress in enabling robots to perform complex manipulation tasks from raw sensory observations. With the rise of large-scale demonstrations~\cite{collaborationOpenXEmbodimentRobotic2023,khazatsky2024droid,walke2023bridgedata} and powerful neural architectures~\cite{vaswani2017attention,hoDenoisingDiffusionProbabilistic2020}, recent methods have increasingly focused on end-to-end learning from visual inputs to low-level control\cite{jang2022bc,brohanRT1RoboticsTransformer2023}.

To better model multi-modal action distributions and mitigate compounding errors, various modeling paradigms have been proposed~\cite{diffusion_policy,act}. For instance, ACT~\cite{act} employs a conditional variational autoencoder to learn action distributions and introduces action chunking to reduce compounding errors. Diffusion Policy~\cite{diffusion_policy} formulates action generation as a denoising process, capturing complex, multi-modal behaviors more effectively. Many subsequent developments have explored enhancements in multiple directions, including enriched sensory inputs~\cite{3d_diffusion_policy,xue2025reactive}, improved network architecture~\cite{flowmatching,robomamba}, expanded datasets\cite{collaborationOpenXEmbodimentRobotic2023}, and scaled model capacity, represented by trend of VLA (vision-language-action) model~\cite{kim24openvla,octo_2023,liu2024rdt,black2024pi_0}.

\begin{wrapfigure}{r}{0.48\linewidth}
    \centering
    \vspace{-10pt}
    \includegraphics[width=\linewidth]{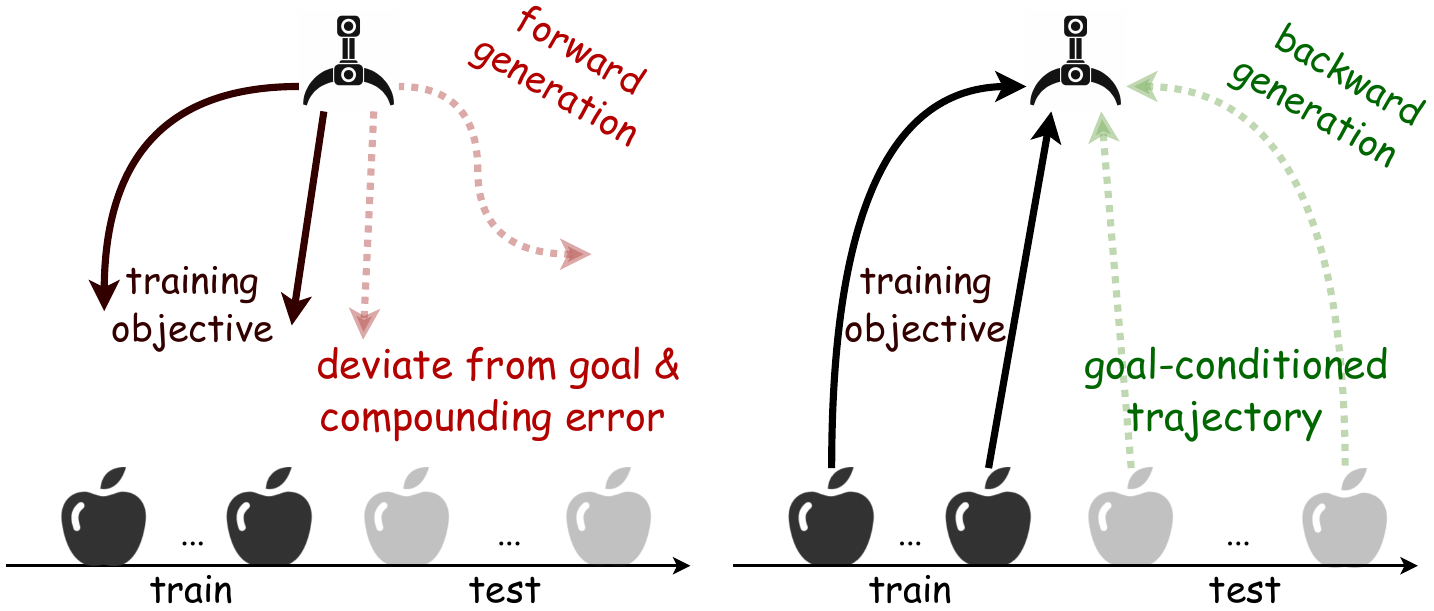}
    \vspace{-10pt}
    \caption{\small{
        \textbf{Comparison between a conventional visuo-motor policy (left) and our proposed Chain-of-Action (right). }
        The former is optimized to predict step-wise actions based on current observations, rather than long-term goals, 
        often leading misaligned behaviors during execution. 
        In contrast, Chain-of-Action adopts a backward generation paradigm, producing goal-conditioned trajectories that reliably execute toward the intended target.
    }}
    \vspace{-10pt}
    \label{fig:teaser}
\end{wrapfigure}

Despite a wide range of these improvements, most of methods still follow a forward prediction paradigm, as illustrated in Figure~\ref{fig:teaser}. While this formulation is intuitive and widely adopted, it suffers from a critical limitation: the accumulation of \emph{compounding errors}~\cite{shi2023waypoint,Kelly2018HGDAggerII,DART,ross2011reduction} during execution. 
The root cause lies in the training objective: these models are optimized to predict the next-step action based on current observation, rather than to ensure successful completion of tasks with long-horizon~\cite{shi2023waypoint}. 
While techniques such as action chunking and image goal conditioned behavioral cloning~\cite{octo_2023,walke2023bridgedata} have been introduced to alleviate compounding errors, they primarily address the symptoms rather than the root cause, which lies in the inherently myopic nature of forward prediction. 


We approach the problem from the \textit{opposite end}, both conceptually and practically, by reversing the action generation process.
While the change in direction may appear simple, it reflects a fundamental shift in how we conceptualize action generation. Instead of predicting actions in a forward, step-wise manner, we propose to construct action sequences in reverse, forming a \emph{chain of actions} that starts from the a keyframe action~\cite{C2F-ARM,shridharPerceiverActorMultiTaskTransformer2022,goyalRVTRoboticView2023a,goyalRVT2LearningPrecise2024}, and  backward towards the initial state. Our insight is that the keyframe action encodes the task-specific goal, which provides a strong structural prior to guide the entire action sequence. By explicitly generating actions from the goal backward, our method enforces a global-to-local consistency~\cite{maHierarchicalDiffusionPolicy2024,xian2023chaineddiffuser} that significantly mitigates compounding errors and enhances generalization under distribution shifts.


To realize this backward reasoning paradigm while maintaining scalability potential~\cite{kaplan2020scaling,tian2024var} for end-to-end training, we unify the entire reverse generation process into a single autoregressive framework. While the formulation is theoretically effective, its practical viability depends on four extra specific designs. These are not optional improvements, but necessary for stable training and reliable closed-loop execution. (1) Continuous action representation:
Discretizing actions into finite bins introduces resolution loss~\cite{openvla-oft,sheebaelhamd2025quantization,liang2025clam}, which becomes particularly problematic in long-horizon autoregressive generation. In our backward generation setup, even small quantization errors can accumulate from the goal backward, leading to significant deviations in earlier steps. To preserve fine-grained structure and trajectory fidelity, we adopt a continuous action representation. (2) Locality action modeling:
While the backward autoregressive structure effectively propagates high-level intent from the goal, it does not explicitly model local action dependencies~\cite{openvla-oft,act,diffusion_policy} within a sub-trajectory. To address this, we adopt a multi-token prediction strategy~\cite{MTP,zhang2025autoregressive} during training, which encourages the model to jointly predict short action chunks. This enhances local coherence and stabilizes training. (3) Dynamic stop:
Closed-loop execution~\cite{mayne1988receding} requires our generation stop at right point. However, in continuous action spaces, there is no discrete end-of-sequence (EOS) token to indicate termination~\cite{zhang2025autoregressive}. We thus design a distance-based stop mechanism that enables the model to determine when to stop based on proximity to the goal, reducing over-generation and improving execution efficiency. (4) Reverse temporal ensemble: Original ensemble strategies~\cite{act}, used in ACT, are designed under forward temporal assumptions and are not directly applicable to our backward generation setting. To address this, we develop a reverse-compatible variant that ensembles multiple backward rollouts, mitigating temporal misalignment and reducing variance during closed-loop execution.

\textbf{Chain-of-Action (CoA)}, which integrates these four essential components into a single autoregressive framework, achieves strong performance in both simulation and real-world settings. CoA outperforms ACT by 16\% and Diffusion Policy by 23\% across 60 RLBench tasks, the most comprehensive evaluation conducted on this benchmark to date, and surpasses ACT by 15\% in real-world robotic manipulation. Crucially, CoA adopts comparable architectures and training setups to ACT, underscoring that the performance gains stem from a principled shift in the modeling paradigm. These results position our trajectory autoregressive modeling as a competitive alternative for visuo-motor policy learning.

%% file: sec/2.related.tex

\label{sec:related}




\para{Hierarchical modeling in robotic manipulation}
A widely adopted strategy in robotic manipulation is to first identify high-level keyframes, and then rely on predefined controllers to handle the low-level execution.  This paradigm is exemplified by C2F-ARM~\cite{C2F-ARM} and extended by methods such as PerAct~\cite{shridharPerceiverActorMultiTaskTransformer2022}, RVT~\cite{goyalRVTRoboticView2023a}, RVT-2~\cite{goyalRVT2LearningPrecise2024}. Recent works like ChainedDiffuser~\cite{xian2023chaineddiffuser} and HDP~\cite{ma2024hierarchical} propose neural planners to replace traditional optimization-based planners. Despite these advances, such methods still operate in an open-loop manner~\cite{xian2023chaineddiffuser,ma2024hierarchical} between keyframes and struggle to adapt to dynamic environments. Our method also builds on the notion of keyframes, but differs fundamentally in its formulation. By unifying keyframe detection and trajectory generation within a single autoregressive framework, it enables efficient environment-aware action prediction and closed-loop execution, where the model can continuously refines its actions based on feedback. As a result, our method no longer relies on high-fidelity 3D inputs for one-shot accurate predictions, which are commonly required by those hierarchical approaches.

\para{CoT-style methods in robotic manipulation}
A separate line of research explores CoT-style VLA agents~\cite{zhao2025cotvla,deng2025graspvla,wen2023any,zheng2024tracevla}, which introduce intermediate semantic representations—such as imagined image goal, visual trace, bounding boxes, or gripper pose, as guidance for subsequent action generation. Orthogonal to these directions, our work focuses on modeling the reasoning process directly between actions without relying on extra modalities as intermediate representations. This design makes our method broadly compatible with different sensory inputs and policy architectures.



%% file: sec/3.2.method.tex
\begin{figure}[t]
    \centering
    \includegraphics[width=\linewidth]{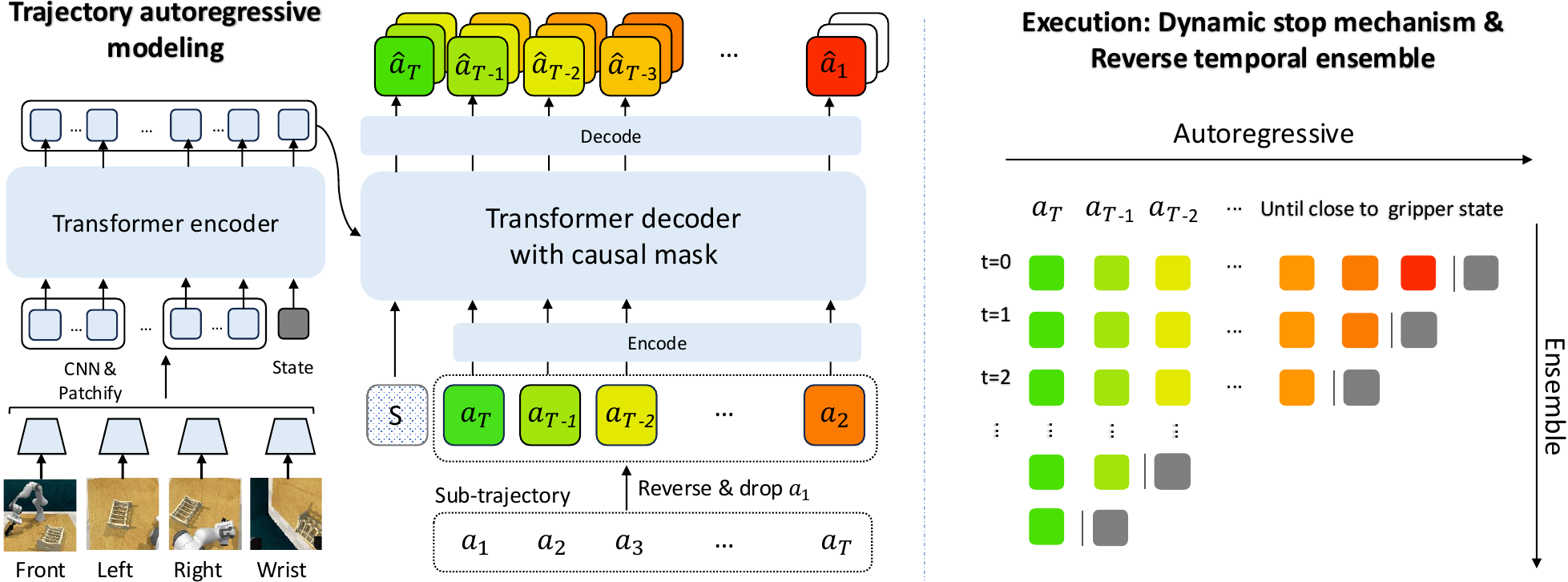}
    \caption{
        \textbf{Chain-of-Action built on trajectory autoregressive modeling.}
        The left part illustrates the network architecture where notation is for the training stage, and the right part illustrates the execution process.
        The model encodes visual and proprioceptive observations and generates actions in reverse order from a predicted keyframe action by an autoregressive decoder.
\textit{For clarity, the keyframe action $a_T$ is shown in green, and subsequent steps are visualized with a gradual color transition.}
    }
    \label{fig:architecture}

\end{figure}

\para{Formulation} \label{sec:modeling} The core idea of Chain-of-Action is to model trajectory generation in reverse: starting from a task-specific goal and predicting actions backward in an autoregressive manner. This reverse formulation imposes a global-to-local structure, anchoring the rollout to the final intent and mitigating compounding errors. An overview of the CoA pipeline is shown on the left of Figure~\ref{fig:architecture}. We adopt the definition of keyframe originally from C2F-ARM~\cite{C2F-ARM}, where a keyframe is identified as a time step at which the gripper state changes or the joint velocities approach zero. This simple yet effective heuristic captures transitions between semantically meaningful phases, such as grasp completion or object placement, and can be interpreted as a task-specific goal. Representing the goal as an action allows it to share the same embedding space with all other actions, enabling seamless backward generation. For each training sample, CoA learns to model the action sequence in reverse order using an autoregressive decoder. \emph{This formulation enforces a reverse causal dependency among actions, yielding a goal-conditioned reasoning chain. Such backward chaining lies at the heart of the our framework, which models the trajectory distribution as:}


\begin{equation}
    p(a_{1:T} \mid O) 
    = 
    \underbrace{p(a_T \mid O)}_{\text{Keyframe Action}}
    \cdot 
    \underbrace{p(a_{T-1} \mid a_T, O) \dots p(a_2 \mid a_{3:T}, O)}_{\text{Reverse Reasoning Actions}}
    \cdot
    \underbrace{p(a_1 \mid a_{2:T}, O)}_{\text{Executed Action}}
\end{equation}

where $\mathbf{a}_T$ denotes the keyframe action, and $\mathbf{O}$ denotes the observation context, including visual input $\mathbf{I}$ and proprioceptive state $\mathbf{S}$. To make the meaning of $\mathbf{a}_{1:T}$ explicit, we clarify how each training sample is constructed. A sub-trajectory is sampled from an expert demonstration by selecting a segment that starts at a random time step and ends at the next first keyframe action. The observation $\mathbf{O}$ is taken from the initial step, and $\mathbf{a}_{1:T}$ denotes the sequence of actions from the current step up to (and including) the keyframe. Each pair $(\mathbf{O}, \mathbf{a}_{1:T})$ forms an independent training example.

  \vspace{10pt}
\begin{minipage}[t]{0.505\linewidth}
  \centering
  \scalebox{0.90}{
  \begin{algorithm}[H]
    \caption{\small{Training Phase}} \label{alg:coa-train}
    \small{
    \textbf{Inputs:} dataset $\mathcal{D}$

    \textbf{Modules:}
    \begin{itemize}[leftmargin=*, noitemsep, topsep=1pt]
      \item Action encoder $f_{\text{enc}}$: $a_t \mapsto x_t$
      \item Action decoder $f_{\text{dec}}$: $x_t \mapsto a_t$
      \item Transformer $F_\theta$: encoder-decoder model
    \end{itemize}

    \textbf{Parameters:} learned token $x_{\text{SOS}}$, loss weight $\lambda$

    \For{iteration $n = 1, 2, \dots$}{
      Sample $(\mathbf{I}, \mathbf{S}, \tau = (a_1, \dots, a_T))$ from $\mathcal{D}$ based on keyframe heuristic

      $x_{1:T} \leftarrow \textsc{Reverse}(f_{\text{enc}}(a_{1:T}))$

      $H \leftarrow \textsc{Concat}(x_{\text{SOS}}, x_{1:T-1})$

      $\hat{x}_{1:T} \leftarrow F_\theta(H \mid \mathbf{I}, \mathbf{S})$

      $\hat{a}_{1:T} \leftarrow \textsc{Reverse}(f_{\text{dec}}(\hat{x}_{1:T}))$

      $\mathcal{L}_{\text{reg}} \leftarrow \sum_{t=1}^{T} \mathcal{L}_{\text{action}}(\hat{a}_t, a_t)$

      $\mathcal{L}_{\text{latent}} \leftarrow \sum_{t=1}^{T} \mathcal{L}_{\text{latent}}(\hat{x}_t, f_{\text{enc}}(a_t))$

      $\mathcal{L}_{\text{total}} \leftarrow \mathcal{L}_{\text{reg}} + \lambda \cdot \mathcal{L}_{\text{latent}}$

      Update $\theta$, $x_{\text{SOS}}$ via backprop on $\mathcal{L}_{\text{total}}$
    }
    }
  \end{algorithm}
  }
\end{minipage}
  \hfill
  \begin{minipage}[t]{0.475\linewidth}
    \centering
    \scalebox{0.90}{
    \begin{algorithm}[H]
      \caption{\small{Inference Phase}} \label{alg:coa-infer}
      \small{
      \textbf{Inputs:} image $\mathbf{I}$, proprioceptive state $\mathbf{S}$

      \textbf{Modules:}
      \begin{itemize}[leftmargin=*, noitemsep, topsep=1pt]
        \item Action encoder $f_{\text{enc}}$: $a_t \mapsto x_t$
        \item Action decoder $f_{\text{dec}}$: $x_t \mapsto a_t$
        \item Transformer $F_\theta$: encoder-decoder model
      \end{itemize}

      \textbf{Parameters:} learned token $x_{\text{SOS}}$, max length $T_{\max}$

      Initialize $H \leftarrow [x_{\text{SOS}}]$

      \For{$t = 1$ to $T_{\max}$}{
        $\hat{x}_t \leftarrow F_\theta(H \mid \mathbf{I}, \mathbf{S})$

        Append $\hat{x}_t$ to $H$

        \If{\textsc{Stop}$(f_{\text{dec}}(\hat{x}_t), \mathbf{S})$}{
          break
        }
      }

      Remove $x_{\text{SOS}}$: $H' \leftarrow H[1:]$

      $\hat{a}_{1:T} \leftarrow \textsc{Reverse}(f_{\text{dec}}(H'))$

      \textbf{Return:} action sequence $\hat{a}_{1:T}$
      }
    \end{algorithm}
    }
  \end{minipage}
  \vspace{10pt}

\para{Continuous action token representation} \label{sec:continuous_action_embedding}
CoA adopts continuous action token representation. However, directly training with continuous latent tokens introduces its own challenge. Unlike discrete token embeddings~\cite{kim24openvla} that are fixed indices supervised by a softmax classifier, our latent actions are generated through a learned encoder.  In this setting, imposing loss directly on the action space fails to constrain the latent space to exhibit temporal consistency during autoregressive decoding. As a result, the latent space lacks meaningful regularization, allowing encoding errors to propagate and amplify through the autoregressive process. To address this, we introduce a latent consistency loss to regularize latent action space: $\mathcal{L}_{\text{consistency}} = \lVert \hat{x}_{t} -  f_{\text{enc}}(\mathbf{a}_{t}) \rVert^2,\quad \text{where }  f_{\text{enc}}(\mathbf{a}_{t}) =  W_{\text{enc}} \mathbf{a}_t + b_{\text{enc}}\,$. Here, $\hat{x}_t$ denotes the predicted latent from the previous timestep, and $f_{\text{enc}}(\mathbf{a}_{t})$ is the encoded latent of the ground-truth action. This loss acts as an inductive bias to align the latent space with temporal dynamics, improving autoregressive generation quality.

\para{Locality modeling} \label{sec:mtp}
Multi token prediction (MTP)~\cite{MTP} can serve as a regularization for action locality modeling. We assign the last $K$ layers of the transformer decoder to produce predictions for different future steps. Concretely, layer $k$ predicts token $\hat{x}_{t+k}$, where 
$k=1,...,K$, making the model aware of the mutual dependencies across the next $K$ steps in a single forward pass. This design introduces temporal locality into the decoding process, enhancing stability in long-horizon generation while remain our global-to-local chain-like structure. Importantly, this regularization is applied only during training and removed at inference. 

\para{Dynamic stop} \label{sec:dynamic_stop}
To enable flexible-length trajectory generation in continuous action space, we introduce a distance-based stop criterion. The core idea is to terminate decoding once the predicted action sufficiently approximates the current execution state, indicating that the backward-generated trajectory has successfully reached the present, as shown in bottom-right in Figure~\ref{fig:architecture}. This stop mechanism is agnostic to the specific action representation and can be readily applied to delta actions or joint-space control by adjusting the reference point accordingly.

\begin{figure}[t]
    \centering
    \includegraphics[width=0.8\linewidth]{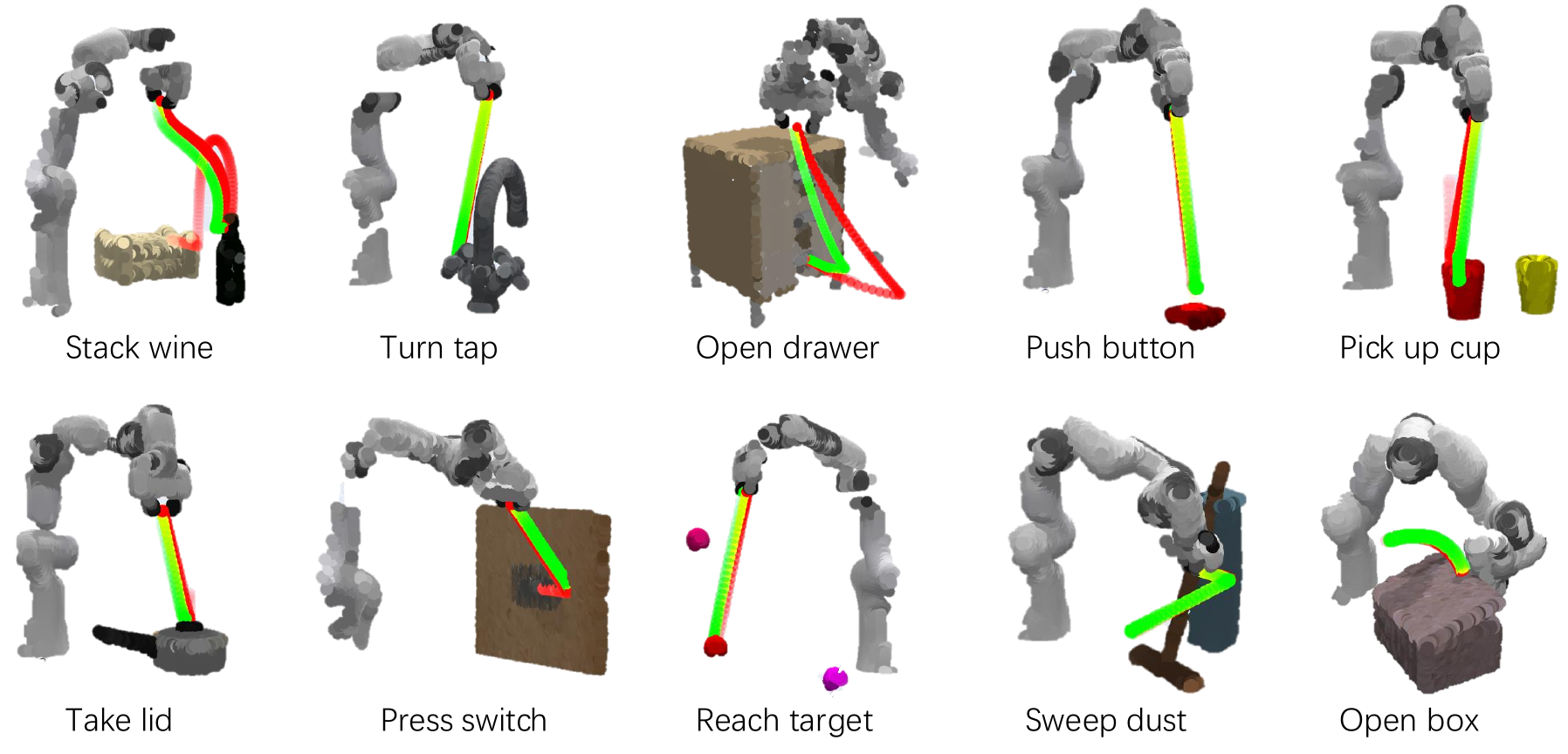}
    \caption{Visualization of predicted sub-trajectories across 10 widely used tasks. Detail refers to Table~\ref{tab:task_success}.
    Red waypoints represent ground-truth trajectories, and green waypoints denote model predictions. 
           Each predicted trajectory is generated backward from a keyframe action to the current gripper state,
        enabling consistent goal-conditioned trajectory generation. The model successfully handles both straight and curved motion patterns.
}
    \label{fig:show_case}

\end{figure}

\para{Reverse temporal ensemble} \label{sec:reverse_ensemble}
We introduce a reverse temporal ensembling strategy tailored for CoA. As shown in the bottom-right corner of Figure~\ref{fig:architecture}, our approach aligns multiple reversed sub-trajectories by their predicted keyframe action $a_T$, which serves as the anchor point for autoregressive decoding. This design offers a unique advantage in CoA: since each trajectory is decoded in reverse from the keyframe, compounding error is inherently constrained by the accuracy of the keyframe action. By further improving the accuracy of the keyframe action through ensembling, we tighten this constraint even more.

%% file: sec/3.3.impl.tex
\vspace{-4pt}
\para{Network architecture}  
Our network follows a similar architecture to ACT~\cite{act}, consisting of a 4-layer Transformer encoder and a 7-layer Transformer decoder. The final decoder layer contains multiple parallel heads for MTP, which are only used during training. 
The observations consist of multi-view RGB images and corresponding states, which are encoded as follows: each image view is processed by a ResNet-18 vision encoder to extract visual tokens.

The gripper state is projected via a learnable linear layer into a token representation. These tokens are then concatenated and passed through the Transformer encoder to produce context features for decoding. Autoregressive action generation is performed by the Transformer decoder, which is initialized with a learnable start-of-sequence (SOS) token corresponding to the final keyframe action. The decoder generates previous actions one step at a time in reverse order, stopping when the predicted action becomes sufficiently close to the current gripper state.  Actions are encoded and decoded into a shared latent embedding space via linear projection layers, which regularized by latent consistency loss as we depicted. Additionally, sinusoidal positional embeddings are added to the action tokens to provide temporal ordering cues.

\para{Training}  
For each training sample, we apply two loss terms: a regression loss in the action space and a consistency loss in the latent space. Both are computed with the MTP regularization, where the model predicts a chunk of $K$ actions at each decoding step. The total loss is defined as:
\begin{equation}
\mathcal{L}_{\text{total}} = \sum_{t=1}^{T} \sum_{k=1}^{K} 
\left\| \hat{\mathbf{a}}_{t+k-1}^{k} - \mathbf{a}_{t+k-1} \right\|^2 
+ \lambda_1 \left\| \hat{x}_{t+k-1}^{k} - f_{\text{enc}}(\mathbf{a}_{t+k-1}) \right\|^2 \,,
\end{equation}
where $\hat{\mathbf{a}}_{t+k-1}^{k}$ and $\hat{x}_{t+k-1}^{k}$ are the predicted action and its latent embedding from $k$-th head of MTP layer at step $t$, and $f_{\text{enc}}(\cdot)$ is the action encoder network. Note that for decoding steps where $t + k - 1 > T$, the corresponding terms are masked out and do not contribute to the loss. This ensures that predictions beyond the trajectory horizon are excluded from supervision. For parallel training with a batch of samples, we set $T_{\text{max}}$ as the maximum sub-trajectory length (practically the longest in the dataset), and zero-pad all shorter sequences accordingly. The loss for padded steps is masked out to avoid affecting gradient updates.

\para{Execution}
For each inference, CoA generates an entire trajectory segment, which can be executed in either open-loop or closed-loop mode. We generally adopt closed-loop control, as it allows reverse temporal ensembling to continuously refine the predicted actions during execution. Under the dynamic stopping setting, we compute the Euclidean distance between the predicted action and the current end-effector pose. This termination criterion is well-suited for our continuous end-effector pose action space.

%% file: sec/5.exp.tex
\begin{wraptable}[14]{r}{0.42\textwidth}
\centering
\small
\setlength{\tabcolsep}{1.6mm}
\renewcommand{\arraystretch}{1.05}
\vspace{-10pt}
\caption{\smallcaption{Success rate across 10 widely-used tasks in RLBench.}}
\label{tab:task_success}
\scalebox{0.93}{
\begin{tabular}{l|cccc}
\toprule
Task & CoA & ACT & DP & Octo\\
\midrule
Stack Wine & \textbf{0.80} & 0.56 & 0.56 &0.52\\
Turn Tap & \textbf{0.56} & 0.36 & 0.32 &0.28\\
Open Drawer & \textbf{0.88} & 0.52 & 0.44 &0.84\\
Push Button & \textbf{0.76} & 0.08 & 0.12 &\textbf{0.76}\\
Pick Up Cup & \textbf{0.80} & 0.20 & 0.00 &0.44\\
Take Lid & \textbf{0.80} & 0.40 & 0.60 & 0.76\\
Press Switch & 0.44 & 0.52 & \textbf{0.56} &0.44\\
Reach Target & 0.84 & \textbf{0.88} & 0.08 &0.60\\
Sweep Dust & 0.92 & \textbf{1.00} & \textbf{1.00} &0.80\\
Open Box & 0.76 & 0.36 & 0.48 &\textbf{0.96}\\
\midrule
Avg. & \textbf{0.756} & 0.488 & 0.416 &0.644\\
\bottomrule
\end{tabular}}
\end{wraptable}

In Sec.~\ref{sec:exp_setting}, we introduce our experiment settings, including simulation environment, train, evaluation settings and metrics. Then we show detailed results of the overall comparison in Section~\ref{sec:ovearall_compa}. To dive into the spatial generalization and obtain better understanding of how CoA work, more specific evaluation is shown in Section~\ref{sec:dive_spa_gener}.  Ablation studies of each components in CoA are shown in Section~\ref{sec:ablation}. Finally, the real-world robot evaluations are shown in Section~\ref{sec:Real-robot Evaluations}

\definecolor{mygray}{gray}{.92}
\vspace{-0.8pt}
\subsection{Simulation experiment settings} \label{sec:exp_setting}
\vspace{-0.8pt}

\para{Simulation setup}
We conduct simulation experiments using RLBench~\cite{James2019RLBenchTR}, a widely-used benchmark built on CoppeliaSim and interfaced via PyRep. The robot is a 7-DoF Franka Emika Panda mounted behind a tabletop workspace. Observations are collected from four RGB cameras (front, left shoulder, right shoulder, and wrist). Images are rendered at a resolution of $128 \times 128$.

\para{Baseline}
We compare our method against representative approaches from three categories:
(1) training visuomotor policies from scratch, including ACT and Diffusion Policy (DP);
(2) finetuned generalist robotic policies, represented by Octo~\cite{octo_2023};
(3) 3D-based hierarchical methods, including PerAct~\cite{shridharPerceiverActorMultiTaskTransformer2022}, 3D Diffuser Actor~\cite{ke20243d}, and RVT-2~\cite{goyalRVT2LearningPrecise2024}. We note 3D-based hierarchical methods fundamentally differ from our approach by relying on 3D inputs and motion planners to generate trajectories. We provide additional discussion on these differences in Appendix~\ref{app:rlbench18}. 

\para{Training and evaluation protocol}
To ensure broad and representative evaluation, our main benchmark is conducted on a tailored set of 60 RLBench tasks, where CoA is compare with ACT and DP, Each method is trained on 100 demonstrations and evluated on 25 demonstrations per task. For ACT and DP, we follow the RLBench training protocol introduced in~\cite{shridharGenerativeImageAction}, which is detailed in Appendix~\ref{app:Hyperparameters}. 
To better demonstrate the effectiveness of our modeling paradigm, we align our base architecture with ACT and introduce modifications primarily in the transformer decoder, as detailed in Section~\ref{sec:Implement}. The strong performance of these baselines—such as perfect success rates on tasks like \textit{Sweep Dust} and competitive results on others (Table~\ref{tab:task_success})—confirms that all reference models are properly trained. For comparison with Octo, we adopt the evaluation subset RLBench-10 proposed in~\cite{zhang2025effective} and use the reported results. This subset is also used for our ablation studies for convenience. To facilitate comparison with 3D-based hierarchical methods, we evaluate on the RLBench-18 split~\cite{shridharPerceiverActorMultiTaskTransformer2022}, using reported results from prior work~\cite{goyalRVT2LearningPrecise2024}.

\begin{figure}[t]
\centering
\begin{tikzpicture}
\node[anchor=south west,inner sep=0] (image) at (0,0) {\includegraphics[width=\linewidth]{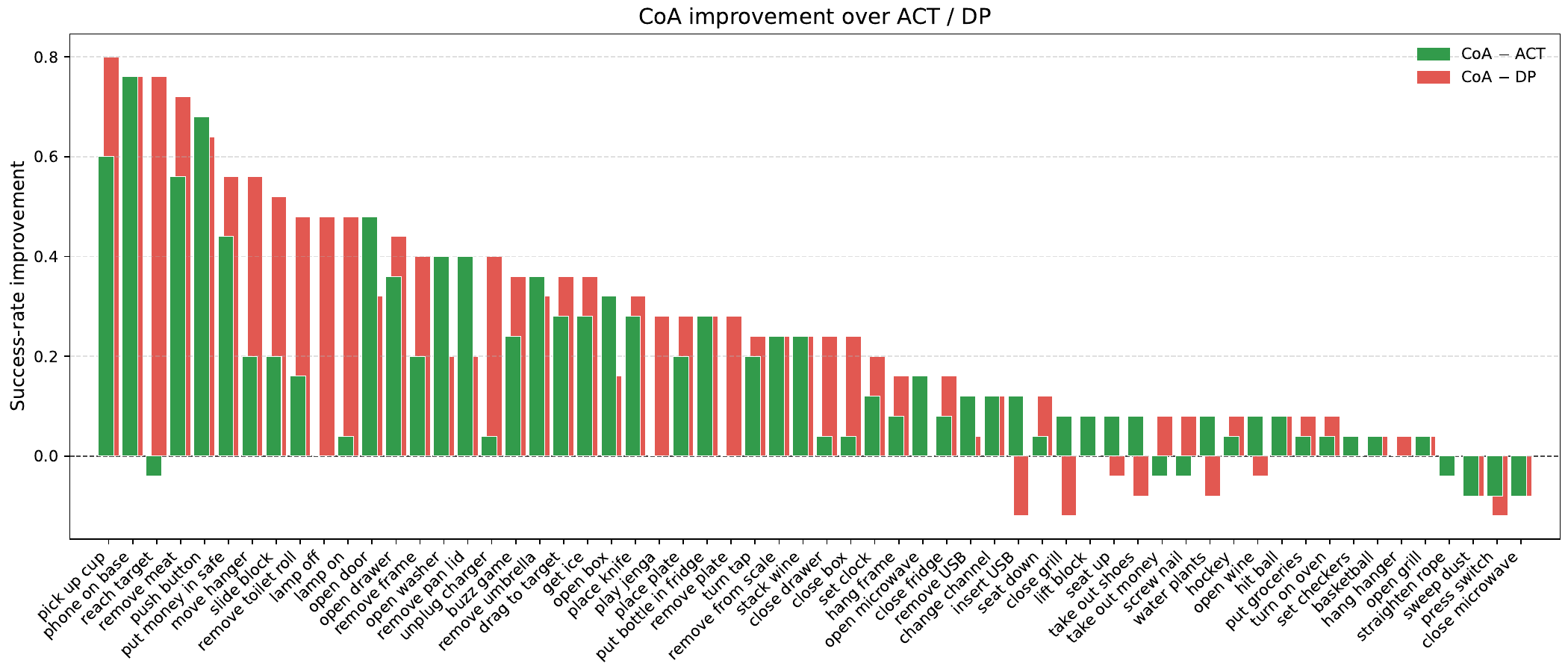}};
\begin{scope}[x={(image.south east)},y={(image.north west)}]

\node[anchor=north west, fill=none, draw=white, font=\scriptsize, inner sep=4pt] at (0.67,0.93) {
\begin{tabular}{lc}

\textbf{Method} & \textbf{Avg. SR} \\
\midrule
ACT & 0.389 \\
DP  & 0.326 \\
CoA & \textbf{0.552} \\

\end{tabular}
};

\end{scope}
\end{tikzpicture}
\vspace{-8pt}
\caption{
Success rate improvement on RLBench-60, sorted by improvement from high to low. 
The average success rate over all tasks is shown in the inset on the right.
}
\label{fig:success_comparison}
\end{figure}

\begin{table}[!htbp]
\centering
\caption{Comparison on the RLBench-18. 3D-based hierarchical methods use 3D point clouds and motion planners, while image-based visuomotor policies operate directly on RGB inputs.}
\vspace{0.2cm}
\label{tab:rlbench18}
\resizebox{\textwidth}{!}{
\begin{tabular}{l|ccc|cccc|c}
\toprule
& \multicolumn{3}{c|}{3D-based hierarchical methods} & \multicolumn{4}{c|}{Image-based visuomotor policies} &  \\
\cmidrule(lr){2-8}
Task & PerAct & 3D Diffuser Actor & RVT-2 & Image-BC (CNN) & Image-BC (ViT) & DP & ACT & CoA \\
\midrule
Close Jar         & 55.2 $\pm$ 4.7 & 96.0 $\pm$ 2.5 & 100.0 $\pm$ 0.0 & 0 & 0 & 0 & 0 & 0 \\
Drag Stick        & 89.6 $\pm$ 4.1 & 100.0 $\pm$ 0.0 & 99.0 $\pm$ 1.7 & 0 & 0 & 0 & 0 & 0 \\
Insert Peg        & 5.6 $\pm$ 4.1 & 65.6 $\pm$ 4.1 & 40.0 $\pm$ 0.0 & 0 & 0 & 0 & 0 & 0 \\
Meat off Grill    & 70.4 $\pm$ 2.0 & 96.8 $\pm$ 1.6 & 99.0 $\pm$ 1.7 & 0 & 0 & 16 & 32 & 88 \\
Open Drawer       & 88.0 $\pm$ 5.7 & 89.6 $\pm$ 4.1 & 74.0 $\pm$ 11.8 & 4 & 0 & 44 & 52 & 88 \\
Place Cups        & 2.4 $\pm$ 3.2 & 24.0 $\pm$ 7.6 & 38.0 $\pm$ 4.5 & 0 & 0 & 0 & 0 & 0 \\
Place Wine        & 44.8 $\pm$ 7.8 & 93.6 $\pm$ 4.8 & 95.0 $\pm$ 3.3 & 0 & 0 & 56 & 56 & 80 \\
Push Buttons      & 92.8 $\pm$ 3.0 & 98.4 $\pm$ 2.0 & 100.0 $\pm$ 0.0 & 0 & 0 & 0 & 32 & 28 \\
Put in Cupboard   & 28.0 $\pm$ 4.4 & 85.6 $\pm$ 4.1 & 66.0 $\pm$ 4.5 & 0 & 0 & 0 & 0 & 8 \\
Put in Drawer     & 51.2 $\pm$ 4.7 & 96.0 $\pm$ 3.6 & 96.0 $\pm$ 0.0 & 8 & 0 & 40 & 60 & 88 \\
Put in Safe       & 84.0 $\pm$ 3.6 & 97.6 $\pm$ 2.0 & 96.0 $\pm$ 2.8 & 4 & 0 & 24 & 36 & 80 \\
Screw Bulb        & 17.6 $\pm$ 2.0 & 82.4 $\pm$ 2.0 & 88.0 $\pm$ 4.9 & 0 & 0 & 0 & 0 & 0 \\
Slide Block       & 74.0 $\pm$ 13.0 & 97.6 $\pm$ 3.2 & 92.0 $\pm$ 2.8 & 0 & 0 & 0 & 36 & 64 \\
Sort Shape        & 16.8 $\pm$ 4.7 & 44.0 $\pm$ 4.4 & 35.0 $\pm$ 7.1 & 0 & 0 & 0 & 0 & 0 \\
Stack Blocks      & 26.4 $\pm$ 3.2 & 68.3 $\pm$ 3.3 & 80.0 $\pm$ 2.8 & 0 & 0 & 0 & 0 & 0 \\
Stack Cups        & 2.4 $\pm$ 2.0 & 47.2 $\pm$ 8.5 & 69.0 $\pm$ 5.9 & 0 & 0 & 0 & 0 & 0 \\
Sweep to Dustpan  & 52.0 $\pm$ 0.0 & 84.0 $\pm$ 4.4 & 100.0 $\pm$ 0.0 & 0 & 0 & 100 & 100 & 92 \\
Turn Tap          & 88.0 $\pm$ 4.4 & 99.2 $\pm$ 1.6 & 99.0 $\pm$ 1.7 & 8 & 16 & 32 & 36 & 56 \\
\midrule
Average  & 48.7 & 81.3 & 81.4 & 1.33 & 0.89 & 17.33 & 24.44 & 37.33 \\
\bottomrule
\end{tabular}}
\end{table}

\subsection{Overall comparisons}
\label{sec:ovearall_compa}
The overall results are presented in Figure~\ref{fig:success_comparison}, with task-wise averages summarized in the accompanying wrapped table. To better assess the effectiveness of our method, we report task-level improvements over both ACT and DP.
Compared to ACT, our method achieves higher success rates on 81.7\% of the tasks, with an average improvement of 16.3\%.
Relative to DP, our method improves performance on 80.0\% of the tasks, with an average gain of 23.2\%. These improvements are especially pronounced in tasks involving significant variation in object position and pose, indicating stronger spatial generalization. As ACT and CoA share a consistent Transformer encoder-decoder architecture and are trained on the same setting, the observed gains highlight the effectiveness of our modeling paradigm. 
The results suggest that a principled change in how action sequences are represented and generated can lead to substantially better performance under distribution shifts. 
The detailed per-task results are in Appendix~\ref{app:rlbench60}. In addition, the comparison with Octo and detailed results with ACT and DP over 10 selected tasks are shown in Table~\ref{tab:task_success}. Results on the RLBench-18 benchmark are reported in Tab~\ref{tab:rlbench18}. We observe that, in these settings, CoA outperforms the finetuned generalist robot policy Octo, while a substantial performance gap remains compared to the 3D-based hierarchical methods.

\subsection{Dive into spatial generalization}
\label{sec:dive_spa_gener}
Although CoA significantly outperforms ACT and DP on the overall benchmark,
it remains crucial to understand \textit{why} such improvements emerge. To this end, we investigate the spatial generalization behavior of our model from three complementary perspectives.

First, in the \textit{Interpolation vs. Extrapolation case study}, we analyze CoA’s performance under controlled spatial distributions within a single representative task.
This study reveals that CoA not only achieves higher success rates under in-distribution (interpolation) configurations, but also demonstrates a substantially larger advantage in out-of-distribution (extrapolation) settings, indicating stronger spatial generalization.

Second, in \textit{Correlation with spatial distribution}, we quantitatively examine how task performance correlates with spatial variation difficulty across the 60 RLBench tasks.
The results show that CoA consistently improves over ACT and DP across all spatial variance levels, and that the performance gap widens as spatial generalization becomes more challenging.

Finally, in \textit{Attention-based analysis of action chain}, we visualize the attention maps between action tokens in the Transformer decoder.
The attention patterns clearly reveal structured dependencies along the predicted action sequence, supporting the hypothesis that CoA performs chain-like global-to-local reasoning throughout the trajectory generation process.

\begin{figure}[t]
\centering
\begin{tikzpicture}

\node[anchor=south west, inner sep=0] (image) at (0,0) {
    \includegraphics[width=0.7\linewidth]{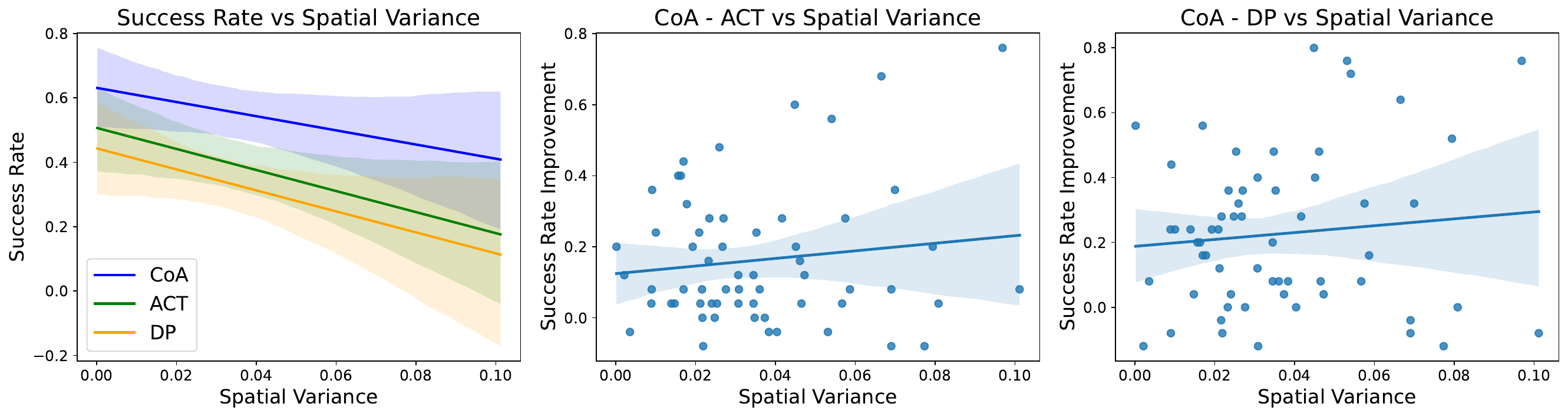}
};
\node[anchor=north west] (table) at ([xshift=10pt,yshift=-4pt]image.north east) {
    \scriptsize
    \scalebox{0.9}{
    \begin{tabular}{lc}
        \toprule
        \textbf{Method/Gap} & \textbf{Pearson ($r$)} \\
        \midrule
        CoA         & -0.1679 \\
        ACT         & -0.2471 \\
        DP          & -0.2455 \\
        CoA - ACT   &  0.1311 \\
        CoA - DP    &  0.1028 \\
        \bottomrule
    \end{tabular}
    }
};

\end{tikzpicture}
\vspace{-4pt}
\caption{\small
\textbf{Correlation between success rate and spatial variance.}
Left image: Overall success rate decreases as object spatial variance increases.
Middle and right image: CoA consistently outperforms ACT and DP across varying spatial generalization levels, with larger advantages in more challenging (higher variance) settings. Table: Pearson correlations highlight CoA’s  robustness to spatial perturbations.}
\label{fig:variance_vs_success_combined}
\vspace{-4pt}
\end{figure}

\para{Correlation with spatial distribution}
We examine the relationship between success rate and the spatial distribution of objects in the evaluation set, aiming to quantify each model’s spatial generalization ability. We use the variance of object coordinates to measure how widely objects are spread in the workspace. As shown in the left plot of Figure~\ref{fig:variance_vs_success_combined}, all methods exhibit a clear trend: success rate decreases as spatial variance increases. This indicates that there spatial generalization becomes more difficult when object placement is more diverse. The improvement plots in the Figure~\ref{fig:variance_vs_success_combined} reveal more details. Compared to ACT and DP, our method consistently outperforms across all levels of spatial variance, and its advantage becomes more pronounced as task difficulty increases.
This trend is further supported by quantitative Pearson correlation.

\begin{figure}[t]
\centering
\begin{tikzpicture}
\coordinate (origin) at (0,0);
\node[anchor=south west, inner sep=0] (image) at (origin) {
    \includegraphics[width=0.70\linewidth]{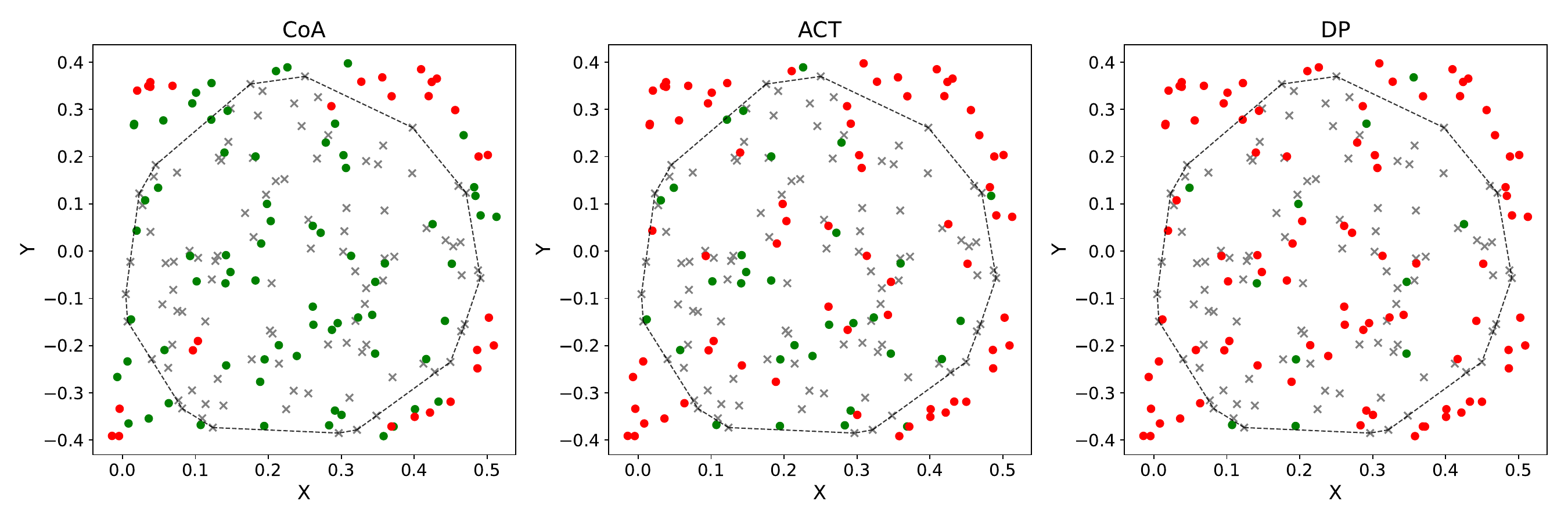}
};
\node[anchor=north west] at ([xshift=10pt,yshift=-20pt]image.north east) {
    \scriptsize
    \renewcommand{\arraystretch}{1.3}  
    \scalebox{0.9}{
    \begin{tabular}{lc}
        \toprule
        \textbf{Method} & \textbf{Inter. / Extra.} \\
        \midrule
        CoA & 0.94 / 0.48 \\
        ACT & 0.54 / 0.08 \\
        DP  & 0.18 / 0.04 \\
        \bottomrule
    \end{tabular}
    }
};

\end{tikzpicture}
\vspace{-4pt}
\caption{\small
\textbf{Interpolate vs. extrapolate performance.}
Success rate comparison on interpolation (in-distribution) and extrapolation (out-of-distribution) subsets for the \textit{Push Button} task. CoA maintains stronger performance across both regimes, with a notably smaller degradation under extrapolation.
}
\label{fig:generalization_plot}
\vspace{-8pt}
\end{figure}

\para{Interpolation vs. Extrapolation case study}
We conduct qualitative analyses on selected tasks to contrast model behavior under interpolated (in-distribution) versus extrapolated (out-of-distribution) spatial configurations.
For this analysis, we choose the \textit{Push Button} task due to its large spatial variation and its frequent use in prior works.
Unlike the standard benchmark setting, we split the dataset into 100 training demonstrations and 100 evaluation demonstrations, further divided into 50 \textit{interpolation} and 50 \textit{extrapolation} samples based on the spatial location of the button.

\begin{figure}[th]
\begin{center}
\includegraphics[width=\linewidth]{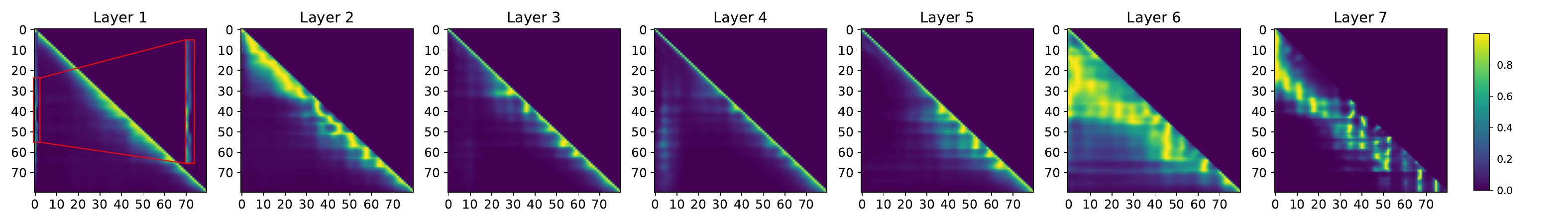}
\caption{\textbf{Attention-based analysis of action chain.} Self-attention maps reveal two key patterns: (1) chain-like dependencies, where each action token attends to recent predecessors, and (2) long-range dependencies (highlighted in the red box in Layer 1), where some tokens directly attend to the initial keyframe action.}
\label{fig:attn_all_layers_row}
\end{center}
\vspace{-4pt}
\label{fig:attn_all_layers}
\vspace{-10pt}
\end{figure}

As shown in Figure~\ref{fig:generalization_plot}, our method outperforms both ACT and DP under both interpolation and extrapolation conditions. Interestingly, while the success rate of CoA in extrapolated settings is about half of that in interpolation, ACT and DP suffer from significantly steeper drops. This highlights the particular difficulty of spatial extrapolation for forward modeling approaches, and suggests that the reverse autoregressive modeling in CoA provides more robust generalization under spatial distribution shifts.

\textbf{Attention-based analysis of action chain}
Figure~\ref{fig:attn_all_layers_row} presents the self-attention maps among action tokens across all decoder layers in our model. The horizontal and vertical indices correspond to the autoregressive decoding order of action tokens, where index 0 denotes keyframe action. We observe two distinct attention patterns: (1) a dominant local chain-like structure, where each action token primarily attends to a recent window of preceding tokens, directly reflecting modeling of CoA; and (2) occasional long-range dependencies (e.g., red box in layer 1 and most of tokens in layer 6), where later tokens exhibit strong attention to initial tokens. This behavior suggests the model leverages the goal-conditioned actions to anchor and guide the full trajectory generation.

\subsection{Ablation on architectural components}
\label{sec:ablation}
We summarize how each architectural component contributes to performance across 10 representative RLBench tasks (selected consistently with Table~\ref{tab:task_success}). The average success rate of each variants are provided in Table~\ref{tab:ablation}.

\para{Modeling paradigm.}
CoA's modeling incorporates two core designs: (1) chain-style autoregressive generation, and (2) goal anchoring via a keyframe action. To assess the necessity of each component, we compare \textit{Reverse} ordering of CoA against two ablated variants:

\textit{Forward} ordering retains the autoregressive structure but removes goal anchoring, starting from the current state and predicting actions forward. Compared to CoA, its lower success rate (0.668 vs. 0.756) highlights the importance of reverse ordering, the core of our proposed modeling. On the other hand, it significantly outperforms ACT (0.668 vs. 0.488), which also uses a autoregressive architecture but predicts fixed-length action chunks. This contrast underscores the advantage of modeling the joint distribution over the entire trajectory, rather than treating it as separated chunks.

\textit{Hybrid} ordering retains goal anchoring but drops chain-style reasoning. It initializes from the keyframe action but switches to forward action generation, removing backward generation process between actions. As a result, the local continuity of autoregressive is lost, and performance drops greatly to 0.600.

\emph{These results confirm that trajectory autoregressive modeling is essential for effective robotic manipulation. Furthermore, reverse autoregressive ordering further enhances performance by anchoring the generation process to the a task-specific goal, providing global guidance throughout the rollout.}

\para{Number of MTP heads}
Multi-token prediction regularization enables the model to capture local action chunks while preserving global causality. Allocating too few heads underutilizes this local context, whereas allocating too many heads disrupts the causal structure. A moderate configuration of 5 heads strikes an effective balance, achieving the highest overall score 0.752.

\begin{wraptable}[20]{r}{0.50\textwidth} 
\centering
\small
\vspace{-10pt}
\caption{Ablation study on individual components by replacing them with alternative settings. The \textbf{bold} indicates the best setting adopted by our final model.}
\label{tab:ablation}
\setlength{\tabcolsep}{4pt}  
\renewcommand{\arraystretch}{1.05}
\begin{tabular}{lcc}
\toprule
\textbf{Components} & \textbf{Setting} & \textbf{Avg. SR} \\
\midrule
\multirow{3}{*}{Modeling Paradigm}
    & Reverse           & \textbf{0.756} \\
    & Forward           & 0.668 \\
    & Hybrid            & 0.600 \\     
    \midrule
\multirow{2}{*}{Embedding Loss}
    & Action consistency & 0.212 \\
 & Latent consistency & \textbf{0.756} \\\midrule
\multirow{2}{*}{Execution}
    &Non-ensemble & 0.66 \\
 & Reverse ensemble & \textbf{0.756} \\\midrule
\multirow{6}{*}{Num. of MTP head}
    & 1   & 0.710 \\
    & 2   & 0.704 \\
    & 4   & 0.720 \\
    & 5   & \textbf{0.756} \\
    & 8   & 0.672 \\
    & 10  & 0.660 \\
\bottomrule
\end{tabular}
\end{wraptable}

\para{Latent consistency loss}
We ablate the latent consistency loss by replacing it with a direct action reconstruction loss, which supervises the action encoder and action decoder to reproduce the input action. This substitution leads to a significant performance drop from 0.752 to 0.212, and results in unstable trajectories with unnatural curling. In contrast, enforcing latent consistency yields a well-structured representation and substantially improves task success.

\para{Reverse temporal ensemble}
We evaluate the impact of reverse temporal ensemble by comparing it with a non-ensemble baseline. Without ensembling, the model achieves 0.660. Applying our reverse-compatible ensemble strategy improves performance to 0.756, highlighting the benefit of aggregating multiple backward rollouts during inference.

\subsection{Real-world experiments}\label{sec:Real-robot Evaluations}
We deploy our method on a Fetch robot featuring a 7-DoF arm and a mobile base for real-world validation. For each task, the robot navigates to a predefined location using its built-in 2D LiDAR-based localization system. Observations are captured from a single RGB camera at $640 \times 480$. resolution and resized to $224 \times 224$. for policy input. Execution is command by absolute end effector poses. To execute commands, we implement a PD controller that calculates the difference between current and desired end effector poses, projects this error into joint space via the Jacobian, and sends velocity commands to the robot. The neural policy operates at 10Hz on a laptop with a 4070 GPU, while the PD controller runs locally on the robot at 1000Hz, with communication handled through ROS for both image data and control commands. 

\begin{figure}[t]
\centering

\begin{minipage}[t]{0.6\linewidth}
    \centering
    \includegraphics[height=4.6cm]{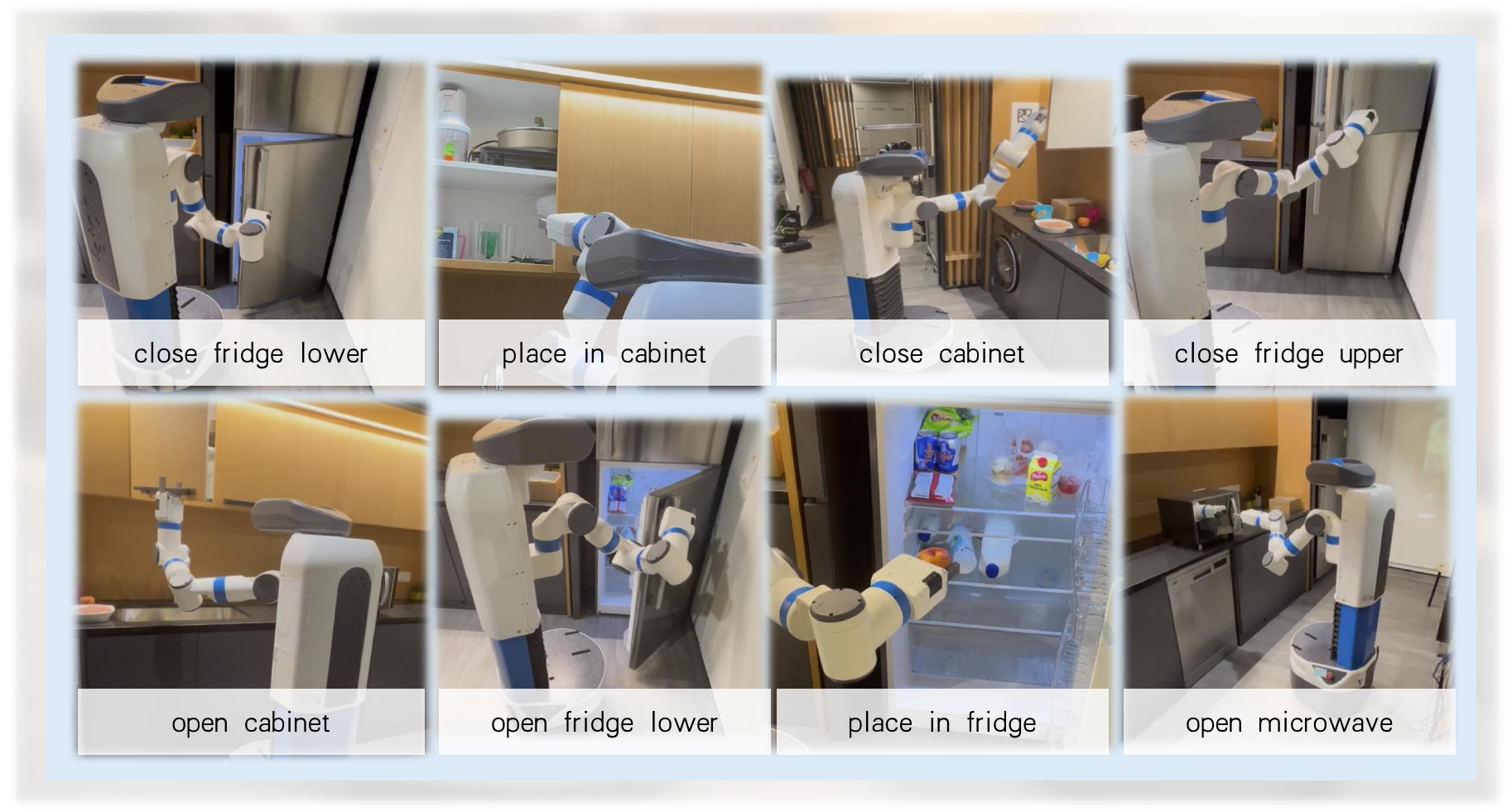}
    \caption{\small{Real-world experiments on 8 kitchen tasks.}}
    \label{fig:real_task_image}
\end{minipage}%
\hfill
\begin{minipage}[t]{0.4\linewidth}
    \centering
    \includegraphics[height=4.6cm]{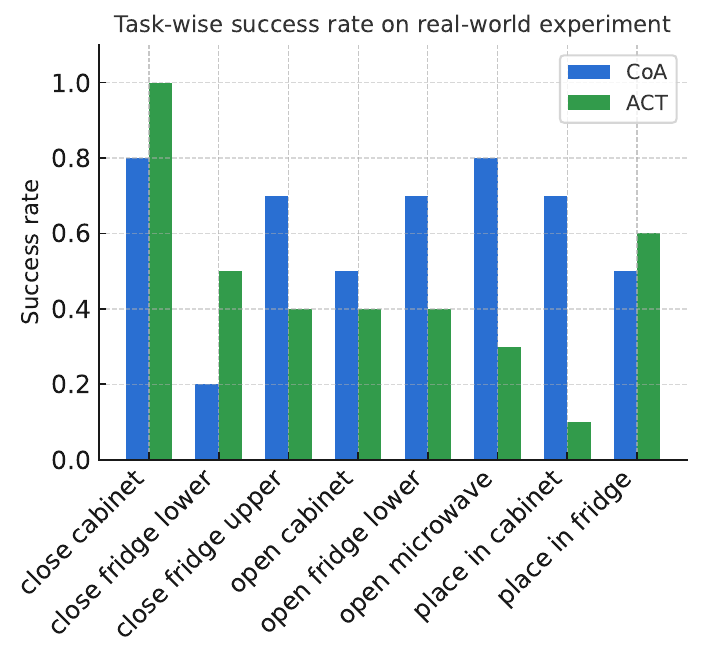}
    \caption{\smallcaption {Real-world experimental results.}}
    \label{fig:real_task_chart}
\end{minipage}
\vspace{-10pt}
\end{figure}

As shown in Figure~\ref{fig:real_task_image}, we evaluated CoA and ACT on 8 kitchen tasks, with the number of expert demonstrations ranging from 35 to 81 per task. Each task was evaluated over 10 trials. The results, summarized in Figure~\ref{fig:real_task_chart}, show that CoA achieves an average success rate of 0.613, outperforming ACT, which achieves 0.463, by a margin of 15\%.

%% file: sec/6.conc.tex
\vspace{-4pt}
We present \textbf{Chain-of-Action}, an action-level reasoning model built upon trajectory autoregressive modeling. By decomposing the joint distribution of the trajectory in reverse, starting from a keyframe and progressing backward to the initial gripper state, our formulation imposes a \textit{global-to-local} structure that enforces consistency between local actions and global task goal. To enable stable training and execution under this backward autoregressive framework, we introduce four necessary design components. Overall, our proposed visuo-motor modeling paradigm significantly improves spatial generalization, and we hope it offers a compelling alternative for future visuomotor policy design. However, the current modeling paradigm relies on keyframe heuristics to split the trajectory, which may not generalize well to diverse task types. Future work can explore learning keyframe structures in an unsupervised manner.

%% file: sec/7.appendix.tex
\section{Comparison on RLBench-18}
\label{app:rlbench18}
The RLBench-18 subset was originally introduced by PerAct~\cite{shridhar2023perceiver} and later became the standard comparison benchmark for 3D hierarchical methods. The results are detailed in Table~\ref{tab:rlbench18}. To clarify the fundamental difference between these two categories of approaches,  
Table~\ref{tab:key_diff_methodology} summarizes their methodological distinctions.  3D-based hierarchical methods typically rely on 3D perception and motion planning, whereas image-based visuomotor policies operate directly on raw RGB observations and learn end-to-end trajectory generation without explicit planners. We observe that, for RGB-only policies, several tasks in RLBench-18 are challenging and frequently result in zero success rates, which limits their discriminative power. This observation motivates our use of the proposed RLBench-60 evaluation split.

\begin{table}[h]
\centering
\caption{Key differences between 3D-based hierarchical methods and image-based visuomotor policies.}
\vspace{0.2cm}
\label{tab:key_diff_methodology}
\resizebox{\textwidth}{!}{
\begin{tabular}{l|c|c}
\toprule
\textbf{Aspect} & \textbf{3D-based Hierarchical Methods} & \textbf{Image-based Visuomotor Policies} \\
\midrule
\textbf{Typical Methods} & PerAct, RVT, RVT-2, 3D Diffuser Actor & ACT, Diffusion Policy, CoA \\
\textbf{Input Modality} & 3D point cloud / RGB-D & RGB-only \\
\textbf{Pipeline} & Two-stage: keyframe action detection + motion planning & End-to-end trajectory prediction without explicit planning \\
\textbf{Execution Mode} & Open-loop execution between keyframes & Closed-loop prediction and control \\
\bottomrule
\end{tabular}}
\end{table}

\section{Per-task success rates on RLBench-60} 
\label{app:rlbench60}

To complement the summary figure in the main paper, which visualizes the performance gap between CoA and baseline methods, we provide the full success rates on all 60 RLBench tasks in Table~\ref{tab:detail_success}. This table lists the per-task success rate of CoA, ACT, and DP, along with the gap of baselines over CoA. Tasks are ordered by the maximum improvement CoA achieves over either baseline, highlighting where our method provides the most substantial gains.

\begin{longtable}{p{2.6cm} p{4.8cm} c c c}
\caption{Detailed results of the overall comparison on RLBench. The simplified names used in Figure~\ref{fig:success_comparison} are matched with their corresponding original task names. The success gap between ACT, DP and CoA is shown as superscripts.}\label{tab:detail_success}\\
\toprule
Simplified name & Original name & CoA & ACT & DP \\
\midrule
\endfirsthead

\toprule
Simplified name & Original name & CoA & ACT & DP \\
\midrule
\endhead
\midrule
\multicolumn{5}{r}{Continued on next page} \\
\midrule
\endfoot

\bottomrule
\endlastfoot
pick up cup & \texttt{pick\_up\_cup} & 0.80 & $0.20^{-0.60}$ & $0.00^{-0.80}$ \\
phone on base & \texttt{phone\_on\_base} & 0.80 & $0.04^{-0.76}$ & $0.04^{-0.76}$ \\
reach target & \texttt{reach\_target} & 0.84 & $0.88^{+0.04}$ & $0.08^{-0.76}$ \\
remove meat & \texttt{meat\_off\_grill} & 0.88 & $0.32^{-0.56}$ & $0.16^{-0.72}$ \\
push button & \texttt{push\_button} & 0.76 & $0.08^{-0.68}$ & $0.12^{-0.64}$ \\
put money in safe & \texttt{put\_money\_in\_safe} & 0.80 & $0.36^{-0.44}$ & $0.24^{-0.56}$ \\
move hanger & \texttt{move\_hanger} & 0.88 & $0.68^{-0.20}$ & $0.32^{-0.56}$ \\
slide block & \texttt{slide\_block\_to\_target} & 0.52 & $0.32^{-0.20}$ & $0.00^{-0.52}$ \\
remove toilet roll & \texttt{take\_toilet\_roll\_off\_stand} & 0.56 & $0.40^{-0.16}$ & $0.08^{-0.48}$ \\
lamp off & \texttt{lamp\_off} & 0.68 & $0.68^{-0.00}$ & $0.20^{-0.48}$ \\
lamp on & \texttt{lamp\_on} & 0.48 & $0.44^{-0.04}$ & $0.00^{-0.48}$ \\
open door & \texttt{open\_door} & 0.92 & $0.44^{-0.48}$ & $0.60^{-0.32}$ \\
open drawer & \texttt{open\_drawer} & 0.88 & $0.52^{-0.36}$ & $0.44^{-0.44}$ \\
remove frame & \texttt{take\_frame\_off\_hanger} & 0.64 & $0.44^{-0.20}$ & $0.24^{-0.40}$ \\
open washer & \texttt{open\_washing\_machine} & 0.76 & $0.44^{-0.32}$ & $0.60^{-0.16}$ \\
remove pan lid & \texttt{take\_lid\_off\_saucepan} & 0.80 & $0.40^{-0.40}$ & $0.60^{-0.20}$ \\
unplug charger & \texttt{unplug\_charger} & 0.60 & $0.56^{-0.04}$ & $0.20^{-0.40}$ \\
buzz game & \texttt{beat\_the\_buzz} & 0.36 & $0.12^{-0.24}$ & $0.00^{-0.36}$ \\
remove umbrella & 
\texttt{take\_\allowbreak umbrella\_\allowbreak out\_\allowbreak of\_\allowbreak umbrella\_\allowbreak stand}
 & 0.52 & $0.16^{-0.36}$ & $0.20^{-0.32}$ \\
drag to target & \texttt{reach\_and\_drag} & 0.64 & $0.36^{-0.28}$ & $0.28^{-0.36}$ \\
get ice & \texttt{get\_ice\_from\_fridge} & 0.60 & $0.32^{-0.28}$ & $0.24^{-0.36}$ \\
open box & \texttt{open\_box} & 0.32 & $0.16^{-0.16}$ & $0.32^{-0.00}$ \\
place knife & \texttt{place\_\allowbreak knife\_\allowbreak on\_\allowbreak chopping\_\allowbreak board} & 0.04 & $0.04^{-0.00}$ & $0.00^{-0.04}$ \\
play jenga & \texttt{play\_\allowbreak jenga} & 1.00 & $1.00^{-0.00}$ & $0.72^{-0.28}$ \\
place plate & \texttt{put\_\allowbreak plate\_\allowbreak in\_\allowbreak colored\_\allowbreak dish\_\allowbreak rack} & 0.32 & $0.12^{-0.20}$ & $0.04^{-0.28}$ \\
put bottle in fridge & \texttt{put\_\allowbreak bottle\_\allowbreak in\_\allowbreak fridge} & 0.28 & $0.00^{-0.28}$ & $0.00^{-0.28}$ \\
remove plate & \texttt{take\_\allowbreak plate\_\allowbreak off\_\allowbreak colored\_\allowbreak dish\_\allowbreak rack} & 0.40 & $0.40^{-0.00}$ & $0.12^{-0.28}$ \\
turn tap & \texttt{turn\_tap} & 0.56 & $0.36^{-0.20}$ & $0.32^{-0.24}$ \\
remove from scale & \texttt{take\_off\_weighing\_scales} & 0.84 & $0.44^{-0.40}$ & $0.64^{-0.20}$ \\
stack wine & \texttt{stack\_wine} & 0.80 & $0.56^{-0.24}$ & $0.56^{-0.24}$ \\
close drawer & \texttt{close\_drawer} & 1.00 & $0.96^{-0.04}$ & $0.76^{-0.24}$ \\
close box & \texttt{close\_box} & 1.00 & $0.96^{-0.04}$ & $0.76^{-0.24}$ \\
set clock & \texttt{change\_clock} & 0.40 & $0.28^{-0.12}$ & $0.20^{-0.20}$ \\
hang frame & \texttt{hang\_frame\_on\_wall} & 0.16 & $0.08^{-0.08}$ & $0.00^{-0.16}$ \\
open microwave & \texttt{open\_microwave} & 0.44 & $0.40^{-0.04}$ & $0.40^{-0.04}$ \\
close fridge & \texttt{close\_fridge} & 0.92 & $0.84^{-0.08}$ & $0.76^{-0.16}$ \\
remove USB & \texttt{take\_usb\_out\_of\_computer} & 0.60 & $0.48^{-0.12}$ & $0.72^{+0.12}$ \\
change channel & \texttt{change\_channel} & 0.12 & $0.00^{-0.12}$ & $0.00^{-0.12}$ \\
insert USB & \texttt{insert\_usb\_in\_computer} & 0.92 & $0.80^{-0.12}$ & $0.88^{-0.04}$ \\
seat down & \texttt{toilet\_seat\_down} & 1.00 & $0.96^{-0.04}$ & $0.88^{-0.12}$ \\
close grill & \texttt{close\_grill} & 0.56 & $0.48^{-0.08}$ & $0.68^{+0.12}$ \\
lift block & \texttt{lift\_numbered\_block} & 0.08 & $0.00^{-0.08}$ & $0.08^{-0.00}$ \\
seat up & \texttt{toilet\_seat\_up} & 0.84 & $0.76^{-0.08}$ & $0.88^{+0.04}$ \\
take out shoes & \texttt{take\_shoes\_out\_of\_box} & 0.08 & $0.00^{-0.08}$ & $0.16^{+0.08}$ \\
take out money & \texttt{take\_money\_out\_safe} & 0.76 & $0.80^{+0.04}$ & $0.68^{-0.08}$ \\
screw nail & \texttt{screw\_nail} & 0.08 & $0.12^{+0.04}$ & $0.00^{-0.08}$ \\
water plants & \texttt{water\_plants} & 0.48 & $0.40^{-0.08}$ & $0.56^{+0.08}$ \\
hockey & \texttt{hockey} & 0.08 & $0.04^{-0.04}$ & $0.00^{-0.08}$ \\
open wine & \texttt{open\_wine\_bottle} & 0.36 & $0.28^{-0.08}$ & $0.40^{+0.04}$ \\
hit ball & \texttt{hit\_ball\_with\_cue} & 0.08 & $0.00^{-0.08}$ & $0.00^{-0.08}$ \\
put groceries & \texttt{put\_groceries\_in\_cupboard} & 0.08 & $0.04^{-0.04}$ & $0.00^{-0.08}$ \\
turn on oven & \texttt{turn\_oven\_on} & 0.36 & $0.32^{-0.04}$ & $0.28^{-0.08}$ \\
set checkers & \texttt{setup\_checkers} & 0.04 & $0.00^{-0.04}$ & $0.04^{-0.00}$ \\
basketball & \texttt{basketball\_in\_hoop} & 0.76 & $0.72^{-0.04}$ & $0.72^{-0.04}$ \\
hang hanger & \texttt{place\_hanger\_on\_rack} & 0.32 & $0.04^{-0.28}$ & $0.00^{-0.32}$ \\
open grill & \texttt{open\_grill} & 0.24 & $0.00^{-0.24}$ & $0.00^{-0.24}$ \\
straighten rope & \texttt{straighten\_rope} & 0.00 & $0.04^{+0.04}$ & $0.00^{-0.00}$ \\
sweep dust & \texttt{sweep\_to\_dustpan} & 0.92 & $1.00^{+0.08}$ & $1.00^{+0.08}$ \\
press switch & \texttt{press\_switch} & 0.44 & $0.52^{+0.08}$ & $0.56^{+0.12}$ \\
close microwave & \texttt{close\_microwave} & 0.72 & $0.80^{+0.08}$ & $0.80^{+0.08}$ \\
\end{longtable}

\section{Supplementary real-world results}

Table~\ref{tab:realworld_sr} reports the per-task success rates of CoA, ACT, and DP across 8 real-world kitchen manipulation tasks. CoA consistently achieves the highest average performance. 


\begin{table}[h]
\centering
\caption{Per-task success rate in real-world experiments.}
\label{app:real-world}
\vspace{0.2cm}
\label{tab:realworld_sr}
\begin{tabular}{lccc}
\toprule
\textbf{Task} & \textbf{CoA} & \textbf{ACT} & \textbf{DP} \\
\midrule
close cabinet       & 0.80 & 1.00 & 0.90 \\
close fridge lower  & 0.20 & 0.50 & 0.60 \\
close fridge upper  & 0.70 & 0.40 & 0.80 \\
open cabinet        & 0.50 & 0.40 & 0.10 \\
open fridge lower   & 0.70 & 0.40 & 0.00 \\
open microwave      & 0.80 & 0.30 & 0.50 \\
place in cabinet    & 0.70 & 0.10 & 0.00 \\
place in fridge     & 0.50 & 0.60 & 0.00 \\
\midrule
Avg.    & \textbf{0.613} & 0.463 & 0.363 \\
\bottomrule
\end{tabular}
\end{table}

\section{Hyperparameters for RLBench} 
\label{app:Hyperparameters}
We provide the training and evaluation hyperparameters for CoA and all baseline methods used in the simulation experiments. To ensure a fair comparison, the hyperparameters for ACT are largely aligned with those of CoA, allowing us to isolate and assess the impact of our proposed modeling paradigm. For DP, we observe slower convergence relative to CoA and ACT, and thus extend its training duration to 100,000 iterations. In addition, we incorporate temporal ensembling into DP following the implementation in ACT. Octo converges substantially faster, and we find that 2,000 training iterations are sufficient. Given that Octo is primarily pretrained on single-camera data, we finetune it using only the front camera, while increasing the image resolution to enhance visual fidelity. All models are trained on a single NVIDIA H100 GPU per task.

\begin{table}[h]
\caption{\small{Hyperparameters for CoA}}
\vspace{0.2cm}
\small
\centering
\setlength{\extrarowheight}{-2pt} 
\begin{tabular}{ll}
\toprule
Backbone & ImageNet-trained ResNet18~\cite{he2016deep} \\[0.05cm]
Action dimension & 8 (3 position + 4 quaternion + 1 gripper) \\[0.05cm]
Cameras & wrist, front, right shoulder, left shoulder \\[0.05cm]
Learning rate & $1e^{-4}$ \\[0.05cm]
Weight decay & $1e^{-4}$ \\[0.05cm]
Image size & $128 \times 128$ \\[0.05cm]
Execution horizon & 1 \\[0.05cm]
Observation horizon & 1 \\[0.05cm]
\# encoder layers & 4 \\[0.05cm]
\# decoder layers & 7 (6 + 1 multi-token prediction layer) \\[0.05cm]
\# heads & 8 \\[0.05cm]
Feedforward dimension & 3200 \\[0.05cm]
Hidden dimension & 512 \\[0.05cm]
Dropout & 0.1 \\[0.05cm]
Iteration & 20000 \\[0.05cm]
Batch size & 128 \\[0.05cm]
Temporal ensembling & true (reverse temporal ensemble) \\[0.05cm]
Action normalization & $[-1, 1]$ \\[0.05cm]
\bottomrule
\end{tabular}
\end{table}

\begin{table}[h]
\caption{\small{Hyperparameters for ACT}}
\small
\centering
\setlength{\extrarowheight}{-2pt} 
\begin{tabular}{ll}
\toprule
Backbone              & ImageNet-trained ResNet18~\cite{he2016deep}                   \\[0.05cm]
Action dimension & 8 (3 position + 4 quaternion + 1 gripper) \\[0.05cm]                              
Cameras         & wrist, front, right shoulder, left shoulder                                                            \\[0.05cm]
Learning rate         & $1e^{-4}$                                                       \\[0.05cm]

Weight decay          & $1e^{-4}$                                                       \\[0.05cm]
Image size        & $128 \times 128$                                                              \\[0.05cm]
Action sequence  & 20                                                              \\[0.05cm]
Execution horizon     & 1
        \\[0.05cm]
Observation horizon & 1 \\[0.05cm]
\# encoder layers     & 4                                                               \\[0.05cm]
\# decoder layers     & 7                                                               \\[0.05cm]
\# heads              & 8                                                               \\[0.05cm]
Feedforward dimension & 3200                                                            \\[0.05cm]
Hidden dimension      & 512                                                             \\[0.05cm]
Dropout               & 0.1                                                             \\[0.05cm]
Iteration                & 20000                                                            \\[0.05cm]
Batch size            & 128   \\[0.05cm]
Temporal ensembling         & true   \\[0.05cm]
Action normalization         & $[-1, 1]$                      \\[0.05cm]
\bottomrule
\end{tabular}
\vspace{0.2cm}
\end{table}

\FloatBarrier                

\begin{table}[ht]
\caption{\small{Hyperparameters for DP}}
\vspace{0.2cm}
\small
\centering
\begin{tabular}{ll}
\toprule
Backbone                  & ImageNet-trained ResNet18 \cite{he2016deep}                     \\[0.05cm]
Noise predictor            & UNet \cite{ronneberger2015u}                                                            \\[0.05cm]
Action dimension      &  8 (3 position + 4 quaternion + 1 gripper)                                              \\[0.05cm]
Cameras         & wrist, front, right shoulder, left shoulder                                                            \\[0.05cm]
Learning rate                  & $1e^{-4}$                                                       \\[0.05cm]
Weight decay                   & $1e^{-6}$                                                       \\[0.05cm]
Image size        & $128 \times 128$                                                              \\[0.05cm]
Observation horizon            & 1                                                               \\[0.05cm]
Action sequence                    & 20                                                              \\[0.05cm]
Execution horizon              & 1                                                          \\[0.05cm]
Train, test diffusion steps & 50, 50                                                          \\[0.05cm]
Hidden dimension               & 512                                                             \\[0.05cm]
Iteration                          & 100000                                                            \\[0.05cm]
Batch size                     & 128                                                             \\[0.05cm]
Temporal ensembling         & true (following ACT's)   \\[0.05cm]
Scheduler  & DDPM \cite{ho2020denoising}                                 
                          \\[0.05cm]
Action normalization         & [-1, 1] \\[0.05cm]

\bottomrule
\end{tabular}
\end{table}

\clearpage  

\begin{table}[ht]
\caption{\small{Hyperparameters for Octo}}
\vspace{0.2cm}
\small
\centering
\begin{tabular}{ll}
\toprule
Pretrained model                  & Octo-small~\cite{octo_2023}                   \\[0.05cm]                           
Action dimension      &  8 (7 delta joints + 1 gripper)                                              \\[0.05cm]
Cameras         &  front                                                            \\[0.05cm]
Learning rate                  & $3e^{-4}$                                                       \\[0.05cm]
Weight decay                   & $1e^{-2}$                                                       \\[0.05cm]
Image size        & $256 \times 256$                                                              \\[0.05cm]
Observation horizon            & 1                                                               \\[0.05cm]
Action sequence                    & 4                                                              \\[0.05cm]
Execution horizon              & 1                                                          \\[0.05cm]
Iteration                          & 2000                                                            \\[0.05cm]
Batch size                     & 128                                                             \\[0.05cm]
Temporal ensembling         & false   \\[0.05cm]
Action normalization         & mean 0, std 1 \\[0.05cm]
Finetuning head & linear head \\[0.05cm]
Image augmentation  & \texttt{resized crop}, \texttt{brightness}, \texttt{contrast}, \texttt{saturation}, \texttt{hue}   \\[0.05cm]

\bottomrule
\end{tabular}

\end{table}

\section{ACT variant with keyframe action}
\label{sec:ablation_keyframe}
To further examine the impact of keyframe action on action sequence modeling, we conduct an additional ablation by modifying the ACT baseline. Specifically, we introduce a variant, ACT+KF, in which an extra keyframe action is appended to ACT's original action chunk.

As shown in Table~\ref{tab:act_kf_comparison}, ACT+KF achieves a higher average success rate (0.516) compared to the original ACT (0.488), indicating that injecting keyframe actions yields marginal improvements. However, the overall gain remains limited.

This result suggests that while keyframe actions may provide some global guidance, they do not substantially improve the final action quality when introduced in this manner. A similar trend is observed in the poor performance of \textit{Hybrid} (Table~\ref{tab:ablation}), a variant of CoA that incorporates both keyframe supervision and causal decoding but lacks trajectory continuity. The limited effectiveness of both ACT+KF and Hybrid underscores a key insight: merely injecting keyframe signals and enforcing an autoregressive structure is not sufficient.  Instead, it is crucial to model the entire trajectory holistically with temporal continuity, whch is explicitly realized in our CoA formulation.

\begin{table}[h]
\centering
\small
\setlength{\tabcolsep}{1.6mm}
\renewcommand{\arraystretch}{1.1}
\caption{Comparison of ACT vs. ACT+KF (with keyframe action) on 10 RLBench tasks.}
\vspace{0.2cm}
\label{tab:act_kf_comparison}
\begin{tabular}{l|cc}
\toprule
\textbf{Task} & \textbf{ACT} & \textbf{ACT+KF} \\
\midrule
Stack Wine            & 0.56 & 0.56 \\
Turn Tap              & 0.36 & 0.32 \\
Open Drawer           & 0.52 & 0.76 \\
Push Button           & 0.08 & 0.16 \\
Pick Up Cup           & 0.20 & 0.36 \\
Take Lid              & 0.40 & 0.40 \\
Press Switch          & 0.52 & 0.28 \\
Reach Target          & 0.88 & 0.72 \\
Sweep Dust            & 1.00 & 0.96 \\
Open Box              & 0.36 & 0.64 \\
\midrule
Avg.         & 0.488 & \textbf{0.516} \\
\bottomrule
\end{tabular}
\end{table}

%% file: paper.bbl
\begin{thebibliography}{48}
\providecommand{\natexlab}[1]{#1}
\providecommand{\url}[1]{\texttt{#1}}
\expandafter\ifx\csname urlstyle\endcsname\relax
  \providecommand{\doi}[1]{doi: #1}\else
  \providecommand{\doi}{doi: \begingroup \urlstyle{rm}\Url}\fi

\bibitem[Black et~al.(2024)Black, Brown, Driess, Esmail, Equi, Finn, Fusai, Groom, Hausman, Ichter, et~al.]{black2024pi_0}
Kevin Black, Noah Brown, Danny Driess, Adnan Esmail, Michael Equi, Chelsea Finn, Niccolo Fusai, Lachy Groom, Karol Hausman, Brian Ichter, et~al.
\newblock A vision-language-action flow model for general robot control.
\newblock \emph{arXiv preprint arXiv:2410.24164}, 2024.

\bibitem[Brohan et~al.(2023)Brohan, Brown, Carbajal, Chebotar, Dabis, Finn, Gopalakrishnan, Hausman, Herzog, Hsu, Ibarz, Ichter, Irpan, Jackson, Jesmonth, Joshi, Julian, Kalashnikov, Kuang, Leal, Lee, Levine, Lu, Malla, Manjunath, Mordatch, Nachum, Parada, Peralta, Perez, Pertsch, Quiambao, Rao, Ryoo, Salazar, Sanketi, Sayed, Singh, Sontakke, Stone, Tan, Tran, Vanhoucke, Vega, Vuong, Xia, Xiao, Xu, Xu, Yu, and Zitkovich]{brohanRT1RoboticsTransformer2023}
Anthony Brohan, Noah Brown, Justice Carbajal, Yevgen Chebotar, Joseph Dabis, Chelsea Finn, Keerthana Gopalakrishnan, Karol Hausman, Alexander Herzog, Jasmine Hsu, Julian Ibarz, Brian Ichter, Alex Irpan, Tomas Jackson, Sally Jesmonth, Nikhil Joshi, Ryan Julian, Dmitry Kalashnikov, Yuheng Kuang, Isabel Leal, Kuang-Huei Lee, Sergey Levine, Yao Lu, Utsav Malla, Deeksha Manjunath, Igor Mordatch, Ofir Nachum, Carolina Parada, Jodilyn Peralta, Emily Perez, Karl Pertsch, Jornell Quiambao, Kanishka Rao, Michael Ryoo, Grecia Salazar, Pannag Sanketi, Kevin Sayed, Jaspiar Singh, Sumedh Sontakke, Austin Stone, Clayton Tan, Huong Tran, Vincent Vanhoucke, Steve Vega, Quan Vuong, Fei Xia, Ted Xiao, Peng Xu, Sichun Xu, Tianhe Yu, and Brianna Zitkovich.
\newblock {{RT-1}}: {{Robotics Transformer}} for {{Real-World Control}} at {{Scale}}.
\newblock In \emph{Robotics: {{Science}} and {{Systems XIX}}}. {Robotics: Science and Systems Foundation}, July 2023.
\newblock ISBN 978-0-9923747-9-2.
\newblock \doi{10.15607/RSS.2023.XIX.025}.

\bibitem[Chi et~al.(2023)Chi, Xu, Feng, Cousineau, Du, Burchfiel, Tedrake, and Song]{diffusion_policy}
Cheng Chi, Zhenjia Xu, Siyuan Feng, Eric Cousineau, Yilun Du, Benjamin Burchfiel, Russ Tedrake, and Shuran Song.
\newblock Diffusion policy: Visuomotor policy learning via action diffusion.
\newblock \emph{The International Journal of Robotics Research}, page 02783649241273668, 2023.

\bibitem[Chisari et~al.(2024)Chisari, Heppert, Argus, Welschehold, Brox, and Valada]{flowmatching}
Eugenio Chisari, Nick Heppert, Max Argus, Tim Welschehold, Thomas Brox, and Abhinav Valada.
\newblock Learning robotic manipulation policies from point clouds with conditional flow matching.
\newblock \emph{arXiv preprint arXiv:2409.07343}, 2024.

\bibitem[Collaboration et~al.(2023)Collaboration, Padalkar, Pooley, Jain, Bewley, Herzog, Irpan, Khazatsky, Rai, Singh, Brohan, Raffin, Wahid, {Burgess-Limerick}, Kim, Sch{\"o}lkopf, Ichter, Lu, Xu, Finn, Xu, Chi, Huang, Chan, Pan, Fu, Devin, Driess, Pathak, Shah, B{\"u}chler, Kalashnikov, Sadigh, Johns, Ceola, Xia, Stulp, Zhou, Sukhatme, Salhotra, Yan, Schiavi, Kahn, Su, Fang, Shi, Amor, Christensen, Furuta, Walke, Fang, Mordatch, Radosavovic, Leal, Liang, {Abou-Chakra}, Kim, Peters, Schneider, Hsu, Bohg, Bingham, Wu, Wu, Luo, Gu, Tan, Oh, Malik, Tompson, Yang, Lim, Silv{\'e}rio, Han, Rao, Pertsch, Hausman, Go, Gopalakrishnan, Goldberg, Byrne, Oslund, Kawaharazuka, Zhang, Rana, Srinivasan, Chen, Pinto, Tan, Ott, Lee, Tomizuka, Du, Ahn, Zhang, Ding, Srirama, Sharma, Kim, Kanazawa, Hansen, Heess, Joshi, Suenderhauf, Di~Palo, Shafiullah, Mees, Kroemer, Sanketi, Wohlhart, Xu, Sermanet, Sundaresan, Vuong, Rafailov, Tian, Doshi, {Mart{\'i}n-Mart{\'i}n}, Mendonca, Shah, Hoque, Julian, Bustamante, Kirmani, Levine,
  Moore, Bahl, Dass, Sonawani, Song, Xu, Haldar, Adebola, Guist, Nasiriany, Schaal, Welker, Tian, Dasari, Belkhale, Osa, Harada, Matsushima, Xiao, Yu, Ding, Davchev, Zhao, Armstrong, Darrell, Jain, Vanhoucke, Zhan, Zhou, Burgard, Chen, Wang, Zhu, Li, Lu, Chebotar, Zhou, Zhu, Xu, Wang, Bisk, Cho, Lee, Cui, Wu, Tang, Zhu, Li, Iwasawa, Matsuo, Xu, and Cui]{collaborationOpenXEmbodimentRobotic2023}
Open X.-Embodiment Collaboration, Abhishek Padalkar, Acorn Pooley, Ajinkya Jain, Alex Bewley, Alex Herzog, Alex Irpan, Alexander Khazatsky, Anant Rai, Anikait Singh, Anthony Brohan, Antonin Raffin, Ayzaan Wahid, Ben {Burgess-Limerick}, Beomjoon Kim, Bernhard Sch{\"o}lkopf, Brian Ichter, Cewu Lu, Charles Xu, Chelsea Finn, Chenfeng Xu, Cheng Chi, Chenguang Huang, Christine Chan, Chuer Pan, Chuyuan Fu, Coline Devin, Danny Driess, Deepak Pathak, Dhruv Shah, Dieter B{\"u}chler, Dmitry Kalashnikov, Dorsa Sadigh, Edward Johns, Federico Ceola, Fei Xia, Freek Stulp, Gaoyue Zhou, Gaurav~S. Sukhatme, Gautam Salhotra, Ge~Yan, Giulio Schiavi, Gregory Kahn, Hao Su, Hao-Shu Fang, Haochen Shi, Heni~Ben Amor, Henrik~I. Christensen, Hiroki Furuta, Homer Walke, Hongjie Fang, Igor Mordatch, Ilija Radosavovic, Isabel Leal, Jacky Liang, Jad {Abou-Chakra}, Jaehyung Kim, Jan Peters, Jan Schneider, Jasmine Hsu, Jeannette Bohg, Jeffrey Bingham, Jiajun Wu, Jialin Wu, Jianlan Luo, Jiayuan Gu, Jie Tan, Jihoon Oh, Jitendra Malik, Jonathan
  Tompson, Jonathan Yang, Joseph~J. Lim, Jo{\~a}o Silv{\'e}rio, Junhyek Han, Kanishka Rao, Karl Pertsch, Karol Hausman, Keegan Go, Keerthana Gopalakrishnan, Ken Goldberg, Kendra Byrne, Kenneth Oslund, Kento Kawaharazuka, Kevin Zhang, Krishan Rana, Krishnan Srinivasan, Lawrence~Yunliang Chen, Lerrel Pinto, Liam Tan, Lionel Ott, Lisa Lee, Masayoshi Tomizuka, Maximilian Du, Michael Ahn, Mingtong Zhang, Mingyu Ding, Mohan~Kumar Srirama, Mohit Sharma, Moo~Jin Kim, Naoaki Kanazawa, Nicklas Hansen, Nicolas Heess, Nikhil~J. Joshi, Niko Suenderhauf, Norman Di~Palo, Nur Muhammad~Mahi Shafiullah, Oier Mees, Oliver Kroemer, Pannag~R. Sanketi, Paul Wohlhart, Peng Xu, Pierre Sermanet, Priya Sundaresan, Quan Vuong, Rafael Rafailov, Ran Tian, Ria Doshi, Roberto {Mart{\'i}n-Mart{\'i}n}, Russell Mendonca, Rutav Shah, Ryan Hoque, Ryan Julian, Samuel Bustamante, Sean Kirmani, Sergey Levine, Sherry Moore, Shikhar Bahl, Shivin Dass, Shubham Sonawani, Shuran Song, Sichun Xu, Siddhant Haldar, Simeon Adebola, Simon Guist, Soroush
  Nasiriany, Stefan Schaal, Stefan Welker, Stephen Tian, Sudeep Dasari, Suneel Belkhale, Takayuki Osa, Tatsuya Harada, Tatsuya Matsushima, Ted Xiao, Tianhe Yu, Tianli Ding, Todor Davchev, Tony~Z. Zhao, Travis Armstrong, Trevor Darrell, Vidhi Jain, Vincent Vanhoucke, Wei Zhan, Wenxuan Zhou, Wolfram Burgard, Xi~Chen, Xiaolong Wang, Xinghao Zhu, Xuanlin Li, Yao Lu, Yevgen Chebotar, Yifan Zhou, Yifeng Zhu, Ying Xu, Yixuan Wang, Yonatan Bisk, Yoonyoung Cho, Youngwoon Lee, Yuchen Cui, Yueh-Hua Wu, Yujin Tang, Yuke Zhu, Yunzhu Li, Yusuke Iwasawa, Yutaka Matsuo, Zhuo Xu, and Zichen~Jeff Cui.
\newblock Open {{X-Embodiment}}: {{Robotic Learning Datasets}} and {{RT-X Models}}, October 2023.

\bibitem[Deng et~al.(2025)Deng, Yan, Wei, Ma, Yang, Chen, Zhang, Yang, Zhang, Cui, et~al.]{deng2025graspvla}
Shengliang Deng, Mi~Yan, Songlin Wei, Haixin Ma, Yuxin Yang, Jiayi Chen, Zhiqi Zhang, Taoyu Yang, Xuheng Zhang, Heming Cui, et~al.
\newblock Graspvla: a grasping foundation model pre-trained on billion-scale synthetic action data.
\newblock \emph{arXiv preprint arXiv:2505.03233}, 2025.

\bibitem[Gloeckle et~al.(2024)Gloeckle, Idrissi, Rozi{\`e}re, Lopez-Paz, and Synnaeve]{MTP}
Fabian Gloeckle, Badr~Youbi Idrissi, Baptiste Rozi{\`e}re, David Lopez-Paz, and Gabriel Synnaeve.
\newblock Better \& faster large language models via multi-token prediction.
\newblock \emph{arXiv preprint arXiv:2404.19737}, 2024.

\bibitem[Goyal et~al.(2023)Goyal, Xu, Guo, Blukis, Chao, and Fox]{goyalRVTRoboticView2023a}
Ankit Goyal, Jie Xu, Yijie Guo, Valts Blukis, Yu-Wei Chao, and Dieter Fox.
\newblock {{RVT}}: {{Robotic View Transformer}} for {{3D Object Manipulation}}, June 2023.

\bibitem[Goyal et~al.(2024)Goyal, Blukis, Xu, Guo, Chao, and Fox]{goyalRVT2LearningPrecise2024}
Ankit Goyal, Valts Blukis, Jie Xu, Yijie Guo, Yu-Wei Chao, and Dieter Fox.
\newblock {{RVT-2}}: {{Learning Precise Manipulation}} from {{Few Demonstrations}}, June 2024.

\bibitem[He et~al.(2016)He, Zhang, Ren, and Sun]{he2016deep}
Kaiming He, Xiangyu Zhang, Shaoqing Ren, and Jian Sun.
\newblock Deep residual learning for image recognition.
\newblock In \emph{Proceedings of the IEEE conference on computer vision and pattern recognition}, pages 770--778, 2016.

\bibitem[Ho et~al.(2020{\natexlab{a}})Ho, Jain, and Abbeel]{ho2020denoising}
Jonathan Ho, Ajay Jain, and Pieter Abbeel.
\newblock Denoising diffusion probabilistic models.
\newblock \emph{Advances in neural information processing systems}, 33:\penalty0 6840--6851, 2020{\natexlab{a}}.

\bibitem[Ho et~al.(2020{\natexlab{b}})Ho, Jain, and Abbeel]{hoDenoisingDiffusionProbabilistic2020}
Jonathan Ho, Ajay Jain, and Pieter Abbeel.
\newblock Denoising {{Diffusion Probabilistic Models}}, December 2020{\natexlab{b}}.

\bibitem[James and Abbeel(2022)]{C2F-ARM}
Stephen James and Pieter Abbeel.
\newblock Coarse-to-fine q-attention with learned path ranking.
\newblock \emph{arXiv preprint arXiv:2204.01571}, 2022.

\bibitem[James et~al.(2019)James, Ma, Arrojo, and Davison]{James2019RLBenchTR}
Stephen James, Z.~Ma, David~Rovick Arrojo, and Andrew~J. Davison.
\newblock Rlbench: The robot learning benchmark \& learning environment.
\newblock \emph{IEEE Robotics and Automation Letters}, 5:\penalty0 3019--3026, 2019.
\newblock URL \url{https://api.semanticscholar.org/CorpusID:202889132}.

\bibitem[Jang et~al.(2022)Jang, Irpan, Khansari, Kappler, Ebert, Lynch, Levine, and Finn]{jang2022bc}
Eric Jang, Alex Irpan, Mohi Khansari, Daniel Kappler, Frederik Ebert, Corey Lynch, Sergey Levine, and Chelsea Finn.
\newblock Bc-z: Zero-shot task generalization with robotic imitation learning.
\newblock In \emph{Conference on Robot Learning}, pages 991--1002. PMLR, 2022.

\bibitem[Kaplan et~al.(2020)Kaplan, McCandlish, Henighan, et~al.]{kaplan2020scaling}
Jared Kaplan, Sam McCandlish, Tom Henighan, et~al.
\newblock Scaling laws for neural language models.
\newblock \emph{arXiv:2001.08361}, 2020.

\bibitem[Ke et~al.(2024)Ke, Gkanatsios, and Fragkiadaki]{ke20243d}
Tsung-Wei Ke, Nikolaos Gkanatsios, and Katerina Fragkiadaki.
\newblock 3d diffuser actor: Policy diffusion with 3d scene representations.
\newblock \emph{arXiv preprint arXiv:2402.10885}, 2024.

\bibitem[Kelly et~al.(2018)Kelly, Sidrane, Driggs-Campbell, and Kochenderfer]{Kelly2018HGDAggerII}
Michael Kelly, Chelsea Sidrane, K.~Driggs-Campbell, and Mykel~J. Kochenderfer.
\newblock Hg-dagger: Interactive imitation learning with human experts.
\newblock \emph{2019 International Conference on Robotics and Automation (ICRA)}, pages 8077--8083, 2018.
\newblock URL \url{https://api.semanticscholar.org/CorpusID:52939433}.

\bibitem[Khazatsky et~al.(2024)Khazatsky, Pertsch, Nair, Balakrishna, Dasari, Karamcheti, Nasiriany, Srirama, Chen, Ellis, Fagan, Hejna, Itkina, Lepert, Ma, Miller, Wu, Belkhale, Dass, Ha, Jain, Lee, Lee, Memmel, Park, Radosavovic, Wang, Zhan, Black, Chi, Hatch, Lin, Lu, Mercat, Rehman, Sanketi, Sharma, Simpson, Vuong, Walke, Wulfe, Xiao, Yang, Yavary, Zhao, Agia, Baijal, Castro, Chen, Chen, Chung, Drake, Foster, Gao, Herrera, Heo, Hsu, Hu, Jackson, Le, Li, Lin, Lin, Ma, Maddukuri, Mirchandani, Morton, Nguyen, O'Neill, Scalise, Seale, Son, Tian, Tran, Wang, Wu, Xie, Yang, Yin, Zhang, Bastani, Berseth, Bohg, Goldberg, Gupta, Gupta, Jayaraman, Lim, Malik, Martín-Martín, Ramamoorthy, Sadigh, Song, Wu, Yip, Zhu, Kollar, Levine, and Finn]{khazatsky2024droid}
Alexander Khazatsky, Karl Pertsch, Suraj Nair, Ashwin Balakrishna, Sudeep Dasari, Siddharth Karamcheti, Soroush Nasiriany, Mohan~Kumar Srirama, Lawrence~Yunliang Chen, Kirsty Ellis, Peter~David Fagan, Joey Hejna, Masha Itkina, Marion Lepert, Yecheng~Jason Ma, Patrick~Tree Miller, Jimmy Wu, Suneel Belkhale, Shivin Dass, Huy Ha, Arhan Jain, Abraham Lee, Youngwoon Lee, Marius Memmel, Sungjae Park, Ilija Radosavovic, Kaiyuan Wang, Albert Zhan, Kevin Black, Cheng Chi, Kyle~Beltran Hatch, Shan Lin, Jingpei Lu, Jean Mercat, Abdul Rehman, Pannag~R Sanketi, Archit Sharma, Cody Simpson, Quan Vuong, Homer~Rich Walke, Blake Wulfe, Ted Xiao, Jonathan~Heewon Yang, Arefeh Yavary, Tony~Z. Zhao, Christopher Agia, Rohan Baijal, Mateo~Guaman Castro, Daphne Chen, Qiuyu Chen, Trinity Chung, Jaimyn Drake, Ethan~Paul Foster, Jensen Gao, David~Antonio Herrera, Minho Heo, Kyle Hsu, Jiaheng Hu, Donovon Jackson, Charlotte Le, Yunshuang Li, Kevin Lin, Roy Lin, Zehan Ma, Abhiram Maddukuri, Suvir Mirchandani, Daniel Morton, Tony Nguyen,
  Abigail O'Neill, Rosario Scalise, Derick Seale, Victor Son, Stephen Tian, Emi Tran, Andrew~E. Wang, Yilin Wu, Annie Xie, Jingyun Yang, Patrick Yin, Yunchu Zhang, Osbert Bastani, Glen Berseth, Jeannette Bohg, Ken Goldberg, Abhinav Gupta, Abhishek Gupta, Dinesh Jayaraman, Joseph~J Lim, Jitendra Malik, Roberto Martín-Martín, Subramanian Ramamoorthy, Dorsa Sadigh, Shuran Song, Jiajun Wu, Michael~C. Yip, Yuke Zhu, Thomas Kollar, Sergey Levine, and Chelsea Finn.
\newblock Droid: A large-scale in-the-wild robot manipulation dataset.
\newblock 2024.

\bibitem[Kim et~al.(2024)Kim, Pertsch, Karamcheti, Xiao, Balakrishna, Nair, Rafailov, Foster, Sanketi, Vuong, Kollar, Burchfiel, Tedrake, Sadigh, Levine, Liang, and Finn]{kim24openvla}
Moo~Jin Kim, Karl Pertsch, Siddharth Karamcheti, Ted Xiao, Ashwin Balakrishna, Suraj Nair, Rafael Rafailov, Ethan~Paul Foster, Pannag~R. Sanketi, Quan Vuong, Thomas Kollar, Benjamin Burchfiel, Russ Tedrake, Dorsa Sadigh, Sergey Levine, Percy Liang, and Chelsea Finn.
\newblock Openvla: An open-source vision-language-action model.
\newblock In \emph{Conference on Robot Learning}, volume 270, pages 2679--2713. {PMLR}, 2024.

\bibitem[Kim et~al.(2025)Kim, Finn, and Liang]{openvla-oft}
Moo~Jin Kim, Chelsea Finn, and Percy Liang.
\newblock Fine-tuning vision-language-action models: Optimizing speed and success.
\newblock \emph{arXiv preprint arXiv:2502.19645}, 2025.

\bibitem[Laskey et~al.(2017)Laskey, Lee, Fox, Dragan, and Goldberg]{DART}
Michael Laskey, Jonathan Lee, Roy Fox, Anca~D. Dragan, and Ken Goldberg.
\newblock Dart: Noise injection for robust imitation learning.
\newblock In \emph{Conference on Robot Learning}, 2017.
\newblock URL \url{https://api.semanticscholar.org/CorpusID:2043463}.

\bibitem[Liang et~al.(2025)Liang, Czempin, Hong, Zhou, Biyik, and Tu]{liang2025clam}
Anthony Liang, Pavel Czempin, Matthew Hong, Yutai Zhou, Erdem Biyik, and Stephen Tu.
\newblock Clam: Continuous latent action models for robot learning from unlabeled demonstrations.
\newblock \emph{arXiv preprint arXiv:2505.04999}, 2025.

\bibitem[Liu et~al.(2024{\natexlab{a}})Liu, Liu, Wang, An, Li, Zhou, Yang, Zhang, Guo, and Zhang]{robomamba}
Jiaming Liu, Mengzhen Liu, Zhenyu Wang, Pengju An, Xiaoqi Li, Kaichen Zhou, Senqiao Yang, Renrui Zhang, Yandong Guo, and Shanghang Zhang.
\newblock Robomamba: Efficient vision-language-action model for robotic reasoning and manipulation.
\newblock \emph{Advances in Neural Information Processing Systems}, 37:\penalty0 40085--40110, 2024{\natexlab{a}}.

\bibitem[Liu et~al.(2024{\natexlab{b}})Liu, Wu, Li, Tan, Chen, Wang, Xu, Su, and Zhu]{liu2024rdt}
Songming Liu, Lingxuan Wu, Bangguo Li, Hengkai Tan, Huayu Chen, Zhengyi Wang, Ke~Xu, Hang Su, and Jun Zhu.
\newblock Rdt-1b: a diffusion foundation model for bimanual manipulation.
\newblock \emph{arXiv preprint arXiv:2410.07864}, 2024{\natexlab{b}}.

\bibitem[Ma et~al.(2024{\natexlab{a}})Ma, Patidar, Haughton, and James]{ma2024hierarchical}
Xiao Ma, Sumit Patidar, Iain Haughton, and Stephen James.
\newblock Hierarchical diffusion policy for kinematics-aware multi-task robotic manipulation.
\newblock In \emph{Proceedings of the IEEE/CVF Conference on Computer Vision and Pattern Recognition}, pages 18081--18090, 2024{\natexlab{a}}.

\bibitem[Ma et~al.(2024{\natexlab{b}})Ma, Patidar, Haughton, and James]{maHierarchicalDiffusionPolicy2024}
Xiao Ma, Sumit Patidar, Iain Haughton, and Stephen James.
\newblock Hierarchical {{Diffusion Policy}} for {{Kinematics-Aware Multi-Task Robotic Manipulation}}, March 2024{\natexlab{b}}.

\bibitem[Mayne and Michalska(1988)]{mayne1988receding}
David~Q Mayne and Hannah Michalska.
\newblock Receding horizon control of nonlinear systems.
\newblock In \emph{Proceedings of the 27th IEEE Conference on Decision and Control}, pages 464--465. IEEE, 1988.

\bibitem[{Octo Model Team} et~al.(2023){Octo Model Team}, Ghosh, Walke, Pertsch, Black, Mees, Dasari, Hejna, Xu, Luo, Kreiman, Tan, Sadigh, Finn, and Levine]{octo_2023}
{Octo Model Team}, Dibya Ghosh, Homer Walke, Karl Pertsch, Kevin Black, Oier Mees, Sudeep Dasari, Joey Hejna, Charles Xu, Jianlan Luo, Tobias Kreiman, {You Liang} Tan, Dorsa Sadigh, Chelsea Finn, and Sergey Levine.
\newblock Octo: An open-source generalist robot policy.
\newblock \url{https://octo-models.github.io}, 2023.

\bibitem[Ronneberger et~al.(2015)Ronneberger, Fischer, and Brox]{ronneberger2015u}
Olaf Ronneberger, Philipp Fischer, and Thomas Brox.
\newblock U-net: Convolutional networks for biomedical image segmentation.
\newblock In \emph{Medical image computing and computer-assisted intervention--MICCAI 2015: 18th international conference, Munich, Germany, October 5-9, 2015, proceedings, part III 18}, pages 234--241. Springer, 2015.

\bibitem[Ross et~al.(2011)Ross, Gordon, and Bagnell]{ross2011reduction}
St{\'e}phane Ross, Geoffrey Gordon, and Drew Bagnell.
\newblock A reduction of imitation learning and structured prediction to no-regret online learning.
\newblock In \emph{Proceedings of the fourteenth international conference on artificial intelligence and statistics}, pages 627--635. JMLR Workshop and Conference Proceedings, 2011.

\bibitem[Sheebaelhamd et~al.(2025)Sheebaelhamd, Tschannen, Muehlebach, and Vernade]{sheebaelhamd2025quantization}
Ziyad Sheebaelhamd, Michael Tschannen, Michael Muehlebach, and Claire Vernade.
\newblock Quantization-free autoregressive action transformer.
\newblock \emph{arXiv preprint arXiv:2503.14259}, 2025.

\bibitem[Shi et~al.(2023)Shi, Sharma, Zhao, and Finn]{shi2023waypoint}
Lucy~Xiaoyang Shi, Archit Sharma, Tony~Z Zhao, and Chelsea Finn.
\newblock Waypoint-based imitation learning for robotic manipulation.
\newblock \emph{arXiv preprint arXiv:2307.14326}, 2023.

\bibitem[Shridhar et~al.()Shridhar, Lo, and James]{shridharGenerativeImageAction}
Mohit Shridhar, Yat~Long Lo, and Stephen James.
\newblock Generative {{Image}} as {{Action Models}}.

\bibitem[Shridhar et~al.(2022)Shridhar, Manuelli, and Fox]{shridharPerceiverActorMultiTaskTransformer2022}
Mohit Shridhar, Lucas Manuelli, and Dieter Fox.
\newblock Perceiver-{{Actor}}: {{A Multi-Task Transformer}} for {{Robotic Manipulation}}, November 2022.

\bibitem[Shridhar et~al.(2023)Shridhar, Manuelli, and Fox]{shridhar2023perceiver}
Mohit Shridhar, Lucas Manuelli, and Dieter Fox.
\newblock Perceiver-actor: A multi-task transformer for robotic manipulation.
\newblock In \emph{Conference on Robot Learning}, pages 785--799. PMLR, 2023.

\bibitem[Tian et~al.(2024)Tian, Jiang, Yuan, Peng, and Wang]{tian2024var}
Keyu Tian, Yi~Jiang, Zehuan Yuan, Bingyue Peng, and Liwei Wang.
\newblock Visual autoregressive modeling: Scalable image generation via next-scale prediction.
\newblock \emph{arXiv:2404.02905}, 2024.

\bibitem[Vaswani(2017)]{vaswani2017attention}
A~Vaswani.
\newblock Attention is all you need.
\newblock \emph{Advances in Neural Information Processing Systems}, 2017.

\bibitem[Walke et~al.(2023)Walke, Black, Lee, Kim, Du, Zheng, Zhao, Hansen-Estruch, Vuong, He, Myers, Fang, Finn, and Levine]{walke2023bridgedata}
Homer Walke, Kevin Black, Abraham Lee, Moo~Jin Kim, Max Du, Chongyi Zheng, Tony Zhao, Philippe Hansen-Estruch, Quan Vuong, Andre He, Vivek Myers, Kuan Fang, Chelsea Finn, and Sergey Levine.
\newblock Bridgedata v2: A dataset for robot learning at scale.
\newblock In \emph{Conference on Robot Learning (CoRL)}, 2023.

\bibitem[Wen et~al.(2023)Wen, Lin, So, Chen, Dou, Gao, and Abbeel]{wen2023any}
Chuan Wen, Xingyu Lin, John So, Kai Chen, Qi~Dou, Yang Gao, and Pieter Abbeel.
\newblock Any-point trajectory modeling for policy learning.
\newblock \emph{arXiv preprint arXiv:2401.00025}, 2023.

\bibitem[Xian and Gkanatsios(2023)]{xian2023chaineddiffuser}
Zhou Xian and Nikolaos Gkanatsios.
\newblock Chaineddiffuser: Unifying trajectory diffusion and keypose prediction for robotic manipulation.
\newblock In \emph{Conference on Robot Learning/Proceedings of Machine Learning Research}. Proceedings of Machine Learning Research, 2023.

\bibitem[Xue et~al.(2025)Xue, Ren, Chen, Zhang, Fang, Gu, Xu, and Lu]{xue2025reactive}
Han Xue, Jieji Ren, Wendi Chen, Gu~Zhang, Yuan Fang, Guoying Gu, Huazhe Xu, and Cewu Lu.
\newblock Reactive diffusion policy: Slow-fast visual-tactile policy learning for contact-rich manipulation.
\newblock In \emph{RSS}, 2025.

\bibitem[Ze et~al.(2024)Ze, Zhang, Zhang, Hu, Wang, and Xu]{3d_diffusion_policy}
Yanjie Ze, Gu~Zhang, Kangning Zhang, Chenyuan Hu, Muhan Wang, and Huazhe Xu.
\newblock 3d diffusion policy: Generalizable visuomotor policy learning via simple 3d representations.
\newblock In \emph{Robotics: Science and Systems}, 2024.

\bibitem[Zhang et~al.(2025{\natexlab{a}})Zhang, Li, Qiao, Huang, Liu, Dayoub, Ma, and Liu]{zhang2025effective}
Wenbo Zhang, Yang Li, Yanyuan Qiao, Siyuan Huang, Jiajun Liu, Feras Dayoub, Xiao Ma, and Lingqiao Liu.
\newblock Effective tuning strategies for generalist robot manipulation policies.
\newblock In \emph{2025 IEEE International Conference on Robotics and Automation (ICRA)}, pages 7255--7262. IEEE, 2025{\natexlab{a}}.

\bibitem[Zhang et~al.(2025{\natexlab{b}})Zhang, Liu, Chang, Schramm, and Boularias]{zhang2025autoregressive}
Xinyu Zhang, Yuhan Liu, Haonan Chang, Liam Schramm, and Abdeslam Boularias.
\newblock Autoregressive action sequence learning for robotic manipulation.
\newblock \emph{IEEE Robotics and Automation Letters}, 2025{\natexlab{b}}.

\bibitem[Zhao et~al.(2025)Zhao, Lu, Kim, Fu, Zhang, Wu, Li, Ma, Han, Finn, et~al.]{zhao2025cotvla}
Qingqing Zhao, Yao Lu, Moo~Jin Kim, Zipeng Fu, Zhuoyang Zhang, Yecheng Wu, Zhaoshuo Li, Qianli Ma, Song Han, Chelsea Finn, et~al.
\newblock Cot-vla: Visual chain-of-thought reasoning for vision-language-action models.
\newblock \emph{arXiv preprint arXiv:2503.22020}, 2025.

\bibitem[Zhao et~al.(2023)Zhao, Kumar, Levine, and Finn]{act}
Tony~Z. Zhao, Vikash Kumar, Sergey Levine, and Chelsea Finn.
\newblock Learning fine-grained bimanual manipulation with low-cost hardware.
\newblock In \emph{Robotics: Science and Systems XIX, Daegu, Republic of Korea, July 10-14, 2023}, 2023.

\bibitem[Zheng et~al.(2024)Zheng, Liang, Huang, Gao, Daum{\'e}~III, Kolobov, Huang, and Yang]{zheng2024tracevla}
Ruijie Zheng, Yongyuan Liang, Shuaiyi Huang, Jianfeng Gao, Hal Daum{\'e}~III, Andrey Kolobov, Furong Huang, and Jianwei Yang.
\newblock Tracevla: Visual trace prompting enhances spatial-temporal awareness for generalist robotic policies.
\newblock \emph{arXiv preprint arXiv:2412.10345}, 2024.

\end{thebibliography}
